%% file: MAIN.tex
\documentclass[letterpaper]{article} 
\usepackage{aaai24}  
\usepackage{times}  
\usepackage{helvet}  
\usepackage{courier}  
\usepackage[hyphens]{url}  
\usepackage{graphicx} 
\usepackage{natbib}  
\usepackage{caption} 
\usepackage{amsmath} 
\usepackage{amsfonts} 
\usepackage[ruled,vlined,linesnumbered]{algorithm2e}
\usepackage{multirow} 
\usepackage{makecell} 
\usepackage{enumitem}

\usepackage[super]{nth}
\usepackage{bm}
\usepackage{array}
\usepackage{makecell}
\usepackage{multirow}
\usepackage{colortbl, xcolor}
\usepackage{tabularx}
\usepackage{graphicx}
\usepackage{caption}
\usepackage{diagbox}
\usepackage{subcaption}
\usepackage{makecell}
\usepackage{url}
\usepackage{enumitem}
\usepackage{float}
\usepackage[flushleft]{threeparttable}
\usepackage{graphicx} 
\usepackage{xcolor}
\usepackage{amsfonts}
\usepackage{listings}

\usepackage{etoolbox}
\usepackage{environ}

\DeclareMathOperator*{\argmax}{arg\,max}

\usepackage{xcolor,pifont}
\definecolor{cycle2}{RGB}{106, 191, 0}
\definecolor{cycle3}{RGB}{191, 0, 0}

\usepackage{graphicx} 

\title{Knowledge Graph Prompting for Multi-Document Question Answering}
\author {
    Yu Wang,\textsuperscript{\rm 1}
    Nedim Lipka,\textsuperscript{\rm 2}
    Ryan A. Rossi,\textsuperscript{\rm 2}
    Alexa Siu,\textsuperscript{\rm 2}
    Ruiyi Zhang,\textsuperscript{\rm 2}
    Tyler Derr\textsuperscript{\rm 1}
}
\affiliations {
    \textsuperscript{\rm 1} Vanderbilt University, Nashville, USA\\
    \textsuperscript{\rm 2} Adobe Research, San Jose, USA\\
    yu.wang.1@vanderbilt.edu, \{lipka, ryrossi, asiu, ruizhang\}@adobe.com, tyler.derr@vanderbilt.edu
}

\usepackage{bibentry}

\begin{document}
\maketitle

\begin{abstract}
The `pre-train, prompt, predict' paradigm of large language models (LLMs) has achieved remarkable success in open-domain question answering (OD-QA). However, few works explore this paradigm in multi-document question answering (MD-QA), a task demanding a thorough understanding of the logical associations among the contents and structures of documents. To fill this crucial gap, we propose a Knowledge Graph Prompting (KGP) method to formulate the right context in prompting LLMs for MD-QA, which consists of a graph construction module and a graph traversal module. For graph construction, we create a knowledge graph (KG) over multiple documents with nodes symbolizing passages or document structures (e.g., pages/tables), and edges denoting the semantic/lexical similarity between passages or document structural relations. For graph traversal, we design an LLM-based graph traversal agent that navigates across nodes and gathers supporting passages assisting LLMs in MD-QA. The constructed graph serves as the global ruler that regulates the transitional space among passages and reduces retrieval latency. Concurrently, the graph traversal agent acts as a local navigator that gathers pertinent context to progressively approach the question and guarantee retrieval quality. Extensive experiments underscore the efficacy of KGP for MD-QA, signifying the potential of leveraging graphs in enhancing the prompt design and retrieval augmented generation for LLMs. Our code: \url{https://github.com/YuWVandy/KG-LLM-MDQA}.
\end{abstract}

\input{Introduction}

\input{Overview}
\input{Method}
\input{Experiment}

\input{related}

\input{conclusion}

\section*{Ethics Statement}
Due to page limitation, the supplementary material and reproducing details are publically available at \url{https://github.com/YuWVandy/KG-LLM-MDQA/blob/main/AAAI24_KGPrompting_Supplementary.pdf}.

\section*{Acknowledgements}
This research is supported by Adobe Research and the National Science Foundation (NSF) under grant number IIS2239881. 

\bibliography{aaai24}

\input{supplement}

\end{document}

%% file: Introduction.tex
\section{Introduction}
Due to the emergence of large language models (LLMs), the `pre-train, prompt, and predict' paradigm has revolutionized natural language processing (NLP) in real-world applications, such as open-domain question answering, fact-checking, and arithmetic reasoning~\cite{chen2017reading, thorne2018fever, asai2019learning, karpukhin2020dense, aly2021feverous, qin2023chatgpt, zou-caragea-2023-jointmatch, liu2023tackling}. However, no significant efforts have investigated this framework in the scenario of multi-documental question answering (MD-QA), which enjoys practical usage in academic research, customer support, and financial/legal inquiries that require deriving insightful analysis from multiple documents~\cite{tessuto2011legal, bolino2016impression}.

\begin{figure}[t!]
    \centering
    \includegraphics[width=0.5\textwidth]{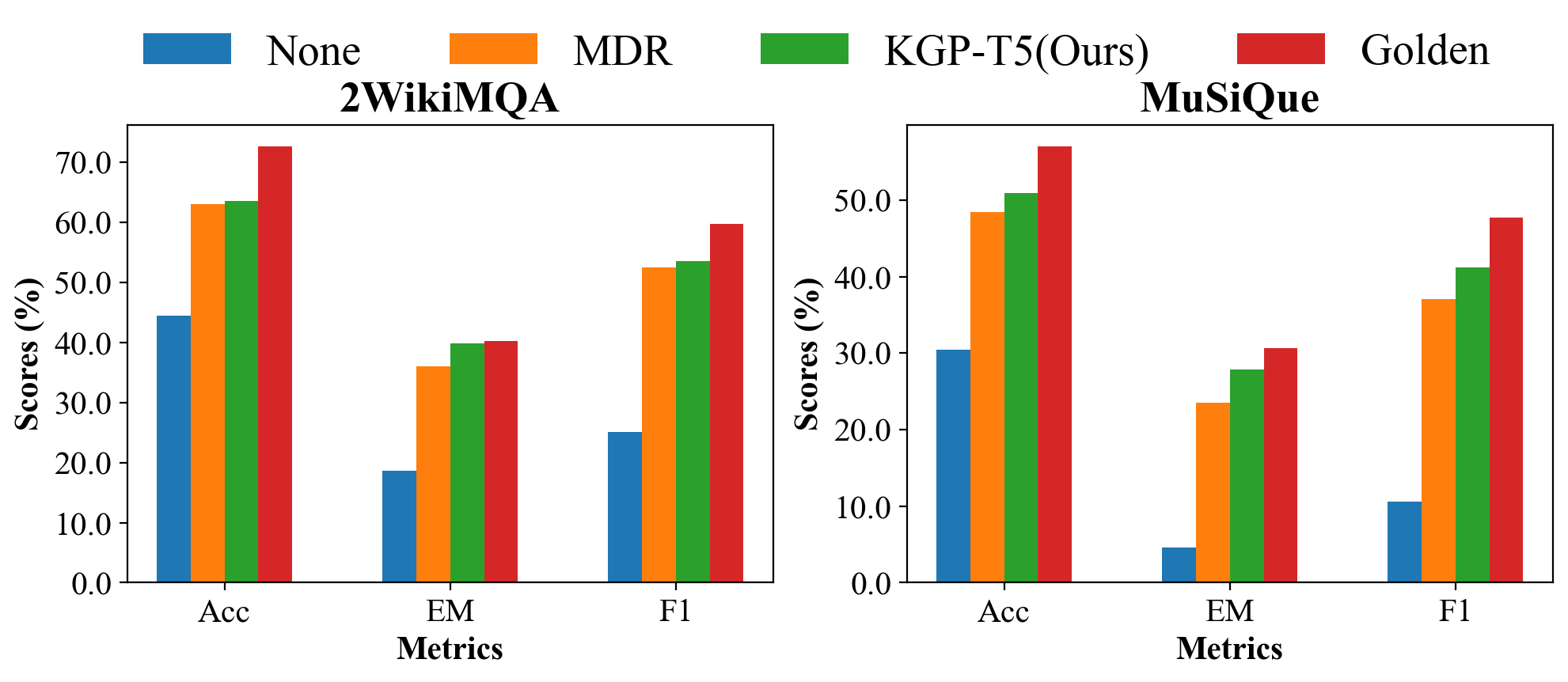}
    \caption{MD-QA performance when prompting ChatGPT with the context retrieved using different strategies.}
    \label{fig-motivate_perform}
    \vspace{-3ex}
\end{figure}

\begin{figure*}[t!]
    \centering
    \includegraphics[width=1\textwidth]{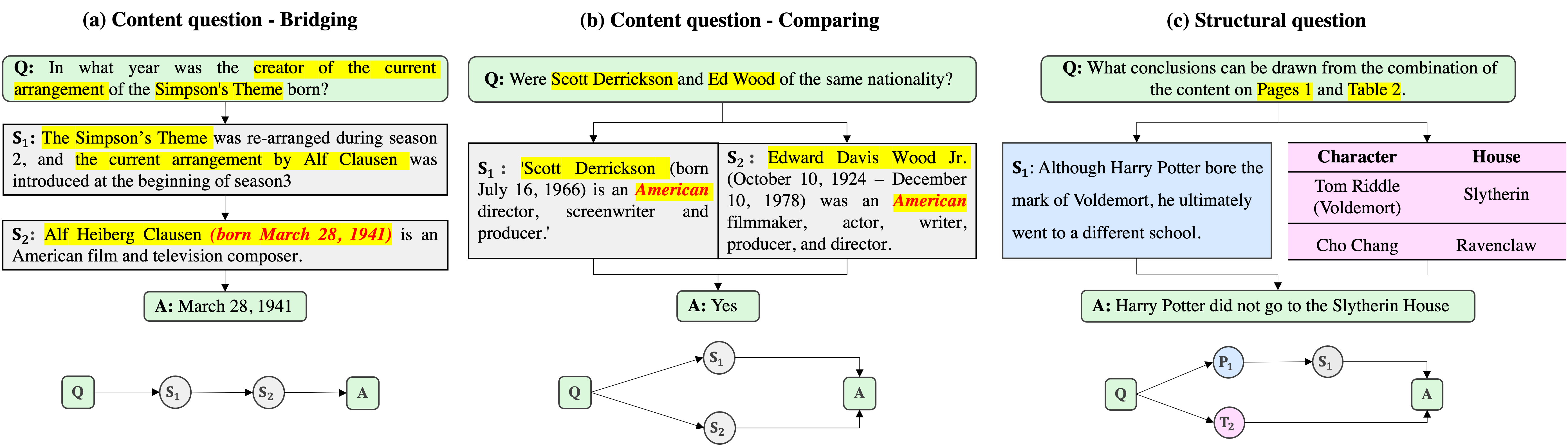}
    \caption{Three popular questions that require reasoning and retrieving over passages/pages/tables from multiple documents. \textbf{(a) Bridging questions} rely on sequential reasoning while \textbf{(b) Comparing questions} rely on parallel reasoning over different passages. \textbf{(c) Structural questions} rely on fetching contents in the corresponding document structures.}
    \label{fig-question}
    \vspace{-2ex}
\end{figure*}

To investigate the capability of LLMs for MD-QA, we randomly sample multi-document questions from the development set of 2WikiMQA~\cite{ho2020constructing} and MuSiQue~\cite{trivedi2022musique}, and then prompt LLMs in four different strategies for the answer\footnote{Detailed experimental setting is presented in Section~\ref{sec-experiment}.}. Successfully answering these questions requires knowledge from multiple Wikipedia documents. As shown in Figure~\ref{fig-motivate_perform}, on 2WikiMQA and MuSiQue, directly prompting LLMs without providing any context, i.e., None, achieves only 25.07\%/10.58\% F1 and 18.60\%/4.60\% EM on 2WikiMQA/MuSiQue, which is far less than 59.69\%/47.75\% F1 and 40.20\%/30.60\% EM when prompting with supporting facts\footnote{Supporting facts: passages that are assumed to contain the answer to the question.} provided as contexts, i.e., the Golden one. This demonstrates the limitation of fulfilling MD-QA using solely the knowledge encoded in LLMs. One common solution to overcome this limitation in conventional OD-QA and single document question-answering (D-QA)~\cite{xu2020layoutlmv2, mathew2021docvqa} is to retrieve grounding contexts and derive faithful answers from the contexts, i.e., retrieve-and-read~\cite{zhu2021retrieving, ju2022grape}. However, unlike OD-QA and D-QA, the primary challenge of MD-QA roots in its demands for alternatively retrieving and reasoning knowledge across different documents~\cite{pereira2023visconde, caciularu2023peek}. For example, successfully answering questions in Figure~\ref{fig-question}(a)-(b)  requires reasoning over distinct passages from two different documents (in these two cases, Wikipedia pages). Moreover, each document is essentially a compilation of multi-modality structured data (e.g., pages, sections, paragraphs, tables, and figures) and some questions may specifically ask for the content in certain structures, which necessitates a comprehensive grasp of these complex document structures. For example, the question in Figure~\ref{fig-question}(c) asks about the difference between Page 1 and Table 2, which is unanswerable if leveraging heuristic methods like BM25 or deep-learning ones like DPR~\cite{karpukhin2020dense}. Building on top of previous challenges, the advent of LLMs introduces new complexities.

For the challenge of alternatively retrieving and reasoning knowledge across different documents, although previous works train a multi-hop retriever~\cite{xiong2020answering, yavuz2022modeling} to imitate such process by sequentially fetching the next passage based on the already-retrieved ones, none of them explore the potential of engaging LLMs into this process. More recent works design different prompting strategies such as Chain/Tree/Graph-of-thought~\cite{trivedi2022interleaving, wei2022chain, yao2023tree, yao2023beyond} to guide LLMs approaching answers progressively. However, prompting non-open-sourced LLMs back and forth incurs forbiddable latency as well as unaffordable consumption. In addition, how to integrate different document structures into the prompt design so that LLMs can understand them is still an open-ended question.

Given the above challenges, we propose a knowledge graph prompting (KGP) method for enhancing LLMs in MD-QA. Specifically, we construct a KG over the given documents with nodes symbolizing passages or document structures and edges denoting their lexical/semantic similarity between passages or intra-document structural relations. Then for the first challenge of alternative reasoning and retrieving knowledge across different documents, we design an LLM-based KG traversal agent, which can alternatively generate the next evidence to approach the question, i.e., reasoning, and select the most promising neighbor to visit from the constructed KG based on the generated evidence, i.e., retrieval. Moreover, we apply the instruction fine-tuning strategy to augment the reasoning capability of the LLM-based KG traversal agent and hence refrain from repeatedly prompting non-open-sourced LLMs for evidence generation. For the multi-modality challenge, we add different types of nodes to the KG characterizing different document structures and hence enabling content retrieval within those specific structures. We highlight our contributions as follows:

\begin{itemize}[leftmargin=*]
    \item \textbf{Generally-applicable KG Construction.} We propose three KG construction methods over documents, with passages or document structures as nodes and their lexical/semantical similarity or structural relations as edges. Then we empirically evaluate the quality of the constructed KGs in MD-QA by checking the level of overlap between the neighborhood and the supporting facts for each question (Figure~\ref{fig-hotpotqa_graph}). Additionally, we provide a comprehensive summary of both our proposed and existing KG construction methods in Table~\ref{tab-compare} in Supplementary.

    \item \textbf{Engaging KG for Prompt Formulation.} We design a Knowledge Graph Prompting (KGP) method, which leverages the LLM-based KG traversal agent to retrieve the question-relevant contexts by traversing the constructed KG. Moreover, we fine-tune this agent to adaptively traverse the most promising neighbors for approaching the question based on the visited nodes (retrieved passages).

    \item \textbf{Case Studies Verifying MD-QA Framework.} We compare the performance of MD-QA when using different types of LLM agents in graph traversal (Table~\ref{tab-ablation}) on the KGs constructed over different numbers of documents (Figure~\ref{fig-analysis}(c)). We conduct case studies on visualizing KGP for MD-QA in Section~\ref{sec-casestudy} in Supplementary.
\end{itemize}

%% file: Overview.tex
\section{Notations}

\begin{figure*}[t]
    \centering
    \includegraphics[width=1.0\textwidth]{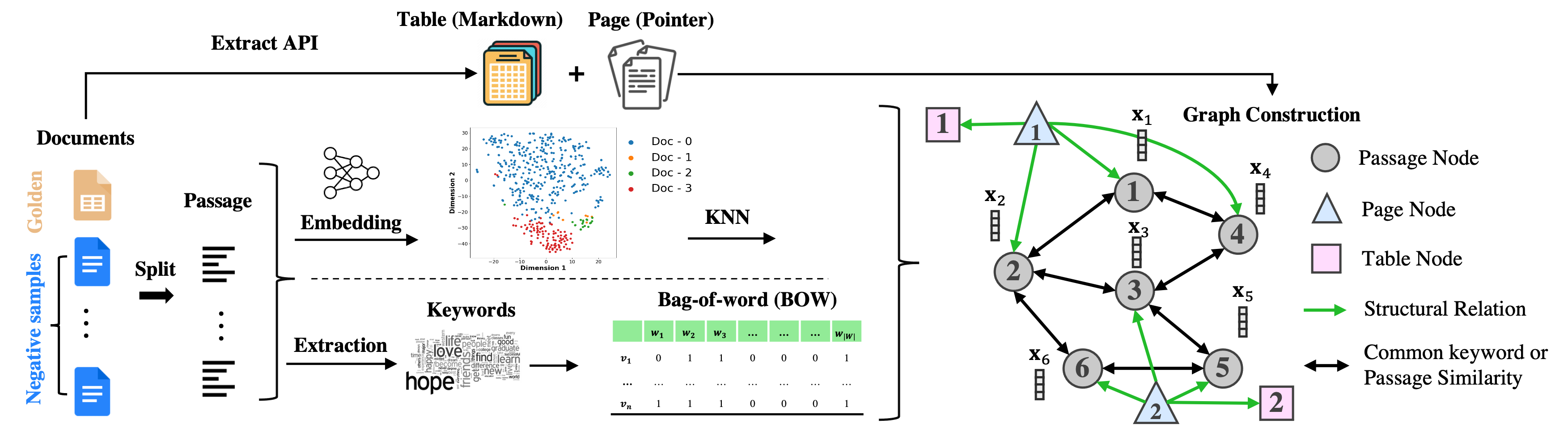}
    \caption{Knowledge Graph Construction. We split each document in the document collection into passages. For each passage, we either directly obtain their embeddings via pre-trained encoders or extract their keywords to build bag-of-word (BOW) features. Then we connect two passages based on their embedding similarity or whether they share common keywords. Additionally, we extract tables/pages via Extract-PDF API and add them as structural nodes to the KG. If pages include passages and tables, we add a directed edge to denote the belonging relations. The table nodes include the markdown formatted content of that table as Figure~\ref{fig-tableunderstanding} in Supplementary has empirically shown that LLMs are able to understand tables in this format.}
    \label{fig-graphconstruct}
    \vspace{-2ex}
\end{figure*}

Following~\cite{tian2023heterogeneous}, let $G = (\mathcal{V}, \mathcal{E})$ be a knowledge graph constructed from a set of documents $\mathcal{D}$, where the node set $\mathcal{V}=\{v_i\}_{i = 1}^{n}$ representing document structures (e.g., passages/pages/tables, etc.) and the edge set $\mathcal{E}\subset \mathcal{V}\times \mathcal{V}$ representing the connections among different nodes (e.g., semantic/lexical similarity and belonging relations among document structures, etc.). Let $\mathcal{X} = \{\mathcal{X}_i\}_i^{n}$ be node features and $\mathcal{X}_i$ corresponds to the feature of node $v_i$, the form of which could be the text for the passage, the markdown for the table and the page number for the page.




%% file: Method.tex
\section{Knowledge Graph Construction}\label{sec-kgconstruct}
Despite numerous well-established KGs~\cite{hoffart2013yago2, tian2023graph}, they treat nodes/edges as entities/relations, which necessitates sophisticated relational extraction techniques and thereby limits their applicability in general domains~\cite{huang2021knowledge}. Additionally, their primary focus on the Wikipedia domain also restricts their usage for answering non-Wikipedia questions such as ones over legal or financial documents. To remedy this issue, we propose generally applicable KG construction methods.

We first analyze two representative questions in Figure~\ref{fig-question}(a)-(b) to motivate our KG construction. Answering these two questions necessitates the deduction of logical associations among different passages. These associations are encoded either through 1) lexical similarity: common keywords shared among different passages, e.g., `Alf Clausen' bridges passage $\mathbf{S}_1$ and passage $\mathbf{S}_2$ in Figure~\ref{fig-question}(a), or 2) semantic similarity: syntactic elements that convey semantic relations, e.g., `nationality' and `American director' in Figure~\ref{fig-question}(b). This motivates us to construct the graph by modeling passages as nodes and their lexical/semantic similarity as edges. More specifically in Figure~\ref{fig-graphconstruct}, we split each document into individual passages, and for each passage $\mathbf{S}_i$, we add a node $v_i$ to the KG with its feature being the text of that passage $\mathcal{X}_i$. Then we add edges by checking the lexical/semantic similarity between pairs of passage nodes. 

\subsubsection{TF-IDF KG Construction} For adding edges according to lexical similarity, we first
apply TF-IDF keyword extraction~\cite{ramos2003using} over each document to filter out meaningless words such as supporting verbs and articles, which also reduces the dimension of bag-of-word (BOW) features, sparsifies the constructed graph and increases the graph traversal efficiency. In addition, we add the document title into the extracted keyword set since some questions focus on title entities. We collect the extracted keywords from all documents to form the keyword space $\mathcal{W}$ and then connect two passages if they share any common keyword in $\mathcal{W}$.


\begin{figure*}[t!]
    \centering
    \includegraphics[width=\textwidth]{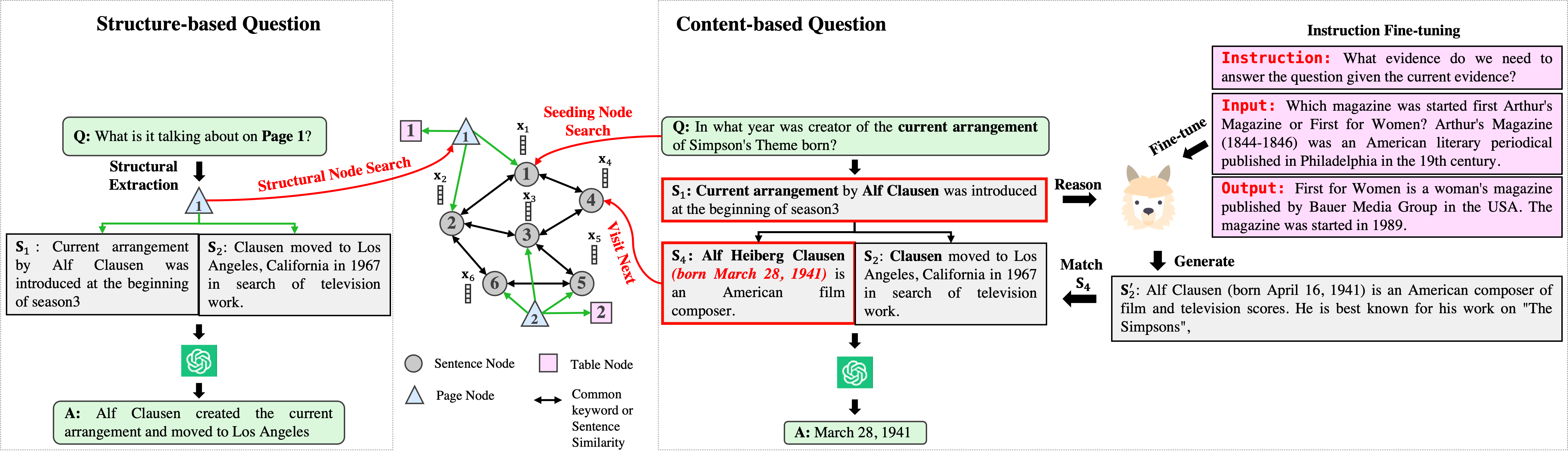}
    \caption{LLM-based KG traversal agent for context retrieval. For questions on document structures (left), we employ LLM to extract structures and retrieve their corresponding contents (the content of pages are passages belonging to that page and the content of tables is the markdown-formatted text). For questions on document content, we concatenate it with the currently retrieved context and prompt the LLM to generate the next evidence to answer the question. By comparing the similarity between the candidate neighboring sentences and the generated passage, we determine the next passage node to traverse. Correspondingly, the candidate neighbors are updated for the next round of traversal.}
    \label{fig-graphwalker}
    \vspace{-2ex}
\end{figure*}

\subsubsection{KNN-ST/MDR KG Construction} For adding edges according to semantic similarity, we can readily employ pre-existing models such as sentence transformers to generate passage embedding $\mathbf{X}_i$ for each node $v_i$ and subsequently compute pairwise similarity matrix to construct the K-nearest neighbor (KNN)
graph. However, these off-the-shelf models, typically trained on tasks not so-related to MD-QA, may not adequately encapsulate necessary logical associations in their embedding similarity demanded by the question. To overcome this problem, we follow the training strategy of MDR~\cite{xiong2020answering} and train a sentence encoder by predicting the subsequent supporting facts based on previously supporting facts, thereby endowing the encoder with reasoning capability. Consequently, the embedding similarity and the corresponding constructed KNN graph fundamentally encapsulate the necessary logical associations between different passages. 

\subsubsection{TAGME} Moreover, we employ TAGME~\cite{min2019knowledge} to extract Wikipedia entities from each passage and construct the graph based on whether two passage nodes share common Wikipedia entities.

In addition to passage nodes, we further add structural nodes into the graph by extracting document structures via Extract-PDF~\footnote{\url{https://developer.adobe.com/document-services/docs/overview/pdf-extract-api/}}. In this paper, we only consider adding pages and tables but the constructed KG can include more different types of document structures. The feature of table nodes is the markdown since LLMs can understand this as demonstrated in Figure~\ref{fig-tableunderstanding} in Supplementary. The feature of page nodes is the page number and we add directed edges from it to sentence/table nodes in that page. \textit{Note that we do not aim to propose a one-size-fits-all KG construction method. Instead, we seek to compare the merits and limitations of various methods in Table~\ref{tab-compare}, offering guidance on which KGs are best suited for specific scenarios.}

\begin{figure}[t!]
    \hspace{-2ex}\includegraphics[width=0.5\textwidth]{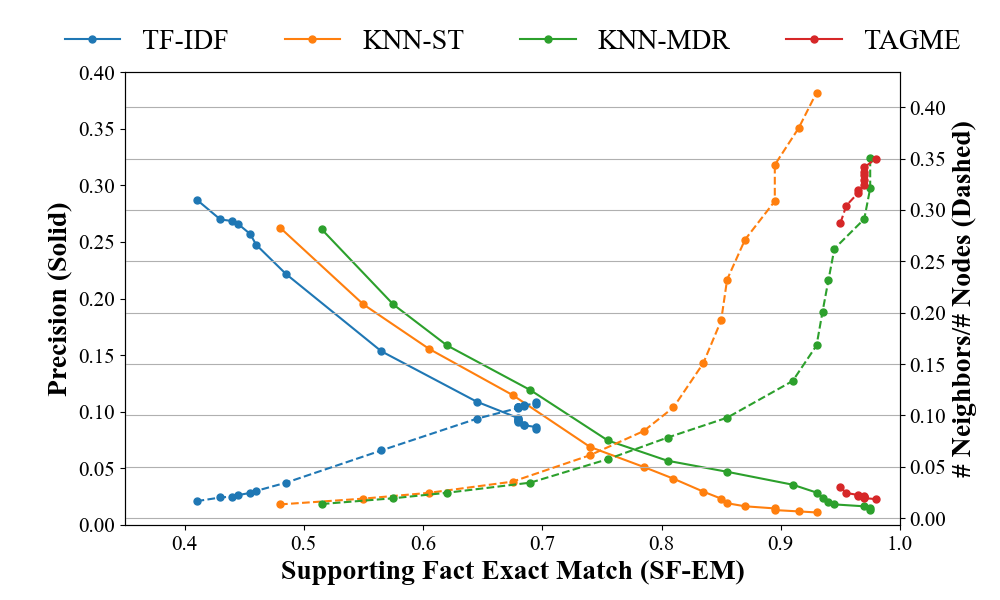}
    \caption{Quality of KGs on HotpotQA. For each KG Construction method, as the average number of neighbors increases (KG becomes denser) in the right y-axis, the SF-EM increases while the precision decreases. KNN-MDR achieves a better trade-off than TF-IDF and KNN-ST. KGs constructed by TAGME are denser than others.}
    \label{fig-hotpotqa_graph}
    \vspace{-5ex}
\end{figure}

\newpage
To verify the constructed KGs indeed encode the necessary information for MD-QA, we randomly sample questions from HotpotQA and construct KGs over the set of documents for each of these questions using our proposed methods. We vary the hyperparameters to control the sparsity of the constructed KG and measure how much percentage of the supporting facts are covered by neighbors of the seeding passages initially searched by TF-IDF. More details about the four construction methods and their hyperparameters are included in Section~\ref{sup-KGC} in Supplementary. As shown in Figure~\ref{fig-hotpotqa_graph}, as the constructed graph becomes denser, the chance that the neighboring node passages hit the supporting facts increases (i.e., SF-EM increases) although the redundant information also increases (i.e., the precision decreases). Given the common keywords shared between one passage to all other passages are typically far less than the total number of passages across all documents, the density of the constructed graph by TF-IDF would be upper-bounded, causing lower SF-EM (evidenced by SF-EM below 0.7 in Figure~\ref{fig-hotpotqa_graph} for TF-IDF curve). For TAGME, we empirically find it identifies a larger quantity of entities mentioned in a single passage, which leads to a denser graph and causes the starting SF-EM of TAGME to be already around 0.95. In addition, since KNN-MDR is pre-trained by predicting the next supporting facts~\cite{xiong2020answering} on HotpotQA, it achieves better trade-off than KNN-ST where the embeddings are directly obtained from the sentence transformer without dataset-specific pre-training.

To summarize, although high SF-EM indicates that the supporting facts for most questions are fully covered by the neighbors of seeding passages, low precision signifies that most of these neighboring passages are irrelevant to the question. Therefore, if we blindly perform graph traversal without any question-tailored adaptation, our retrieved contexts would include redundant passages and compromise the capability of LLMs in MD-QA (which is also verified by the lower performance of KGP-Random in Table~\ref{tab-ablation}). To remedy this issue, in the next section, we introduce an LLM-based KG traversal agent to adaptively visit neighboring passages that are most conducive to answering the given question.

\section{LLM-based KG Traversal Agent}\label{sec-graphwalker}
A natural solution to enable adaptive knowledge graph traversal is to rank the candidate nodes, i.e., the neighbors of the already-visited nodes in our case, thereby determining which ones to visit next. The most straightforward way is to apply heuristic-based fuzzy matching or embedding-based similarity ranking, which cannot capture the intrinsic logical relations between the already traversed paths and the nodes to visit next. Instead, we fine-tune a large language model (LLM) to guide the knowledge graph traversal towards the next most promising passages in approaching the question based on the visited passages, which we term as the LLM-based KG traversal agent.

Given a question $q$ asking about the document content, the LLM-based graph traversal agent reasons over previously visited nodes/retrieved passages $\{s_k\}_{k = 0}^j$ and then generates the next passage $s_{j + 1}$ as follows:
\begin{equation}\label{eq-lmtraversal}
    s_{j + 1} = \argmax_{v \in \mathcal{N}_{j}}
    \phi(g(\mathcal{X}_v), f(||_{k = 0}^{j}\mathcal{X}_k)),
\end{equation}
where $||_{k = 0}^{j}\mathcal{X}_k$ concatenates the textual information of previously retrieved passages/visited nodes. For the choice of $f$, one way is to employ encoder-only models like Roberta-base~\cite{asai2019learning, xiong2020answering, yavuz2022modeling} and correspondingly $g$ would be another encoder model with $\phi(\cdot)$ being the inner product measuring the embedding similarity. Another way is to employ encoder-decoder models such as T5~\cite{brown2020language, touvron2023llama} and correspondingly $g$ would be an identity function with $\phi(\cdot)$ measuring the textual similarity. To mitigate the hallucination issue and enhance the reasoning capability~\cite{wei2022chain, ji2023survey} of the LLM traversal agent, we further instruction fine-tune $f$~\cite{chung2022scaling} by predicting the next supporting facts based on previous supporting facts, thereby integrating commonsense knowledge encoded originally in their pre-trained parameters with the enhanced reasoning capability inherited from the instruction fine-tuning. After visiting the top-scoring nodes selected from the candidate neighbor queue by Eq~$\eqref{eq-lmtraversal}$, the candidate neighbor queue is updated by adding neighbors of these newly visited nodes. We iteratively apply this process until hit the preset budget. Next, we illustrate the above process with an example in Figure~\ref{fig-graphwalker} and present the algorithm thereafter.

Figure~\ref{fig-graphwalker} presents the content-based question asking `In what year was the creator of the current arrangement of Simpson's Theme born?'. We use TF-IDF search to initialize the seeding passage Node 1, which reads: `Alf Heiberg Clausen (born March 28, 1941) is an American film composer'. Subsequently, we prefix the currently retrieved-context (Node 1) with the question and prompt the LLM to generate the next evidence required to approach the question one step closer. Because we augment the reasoning capability of the LLM by instruction fine-tuning, it is expected to recognize the logical association between the question and the currently retrieved context. Consequently, it can predict the subsequent passage that \textit{maintains logical coherence, albeit may contain factual mistakes}, i.e., `Alf Clausen (born April 16, 1941) is an American composer of film and television scores.' To rectify this potential factual mistake, we select nodes from the candidate neighbors that match the most with the LLM-generated passage, in this case, Node 4 `Alf Heiberg Clausen (born March 28, 1941) is an American film composer'. Since this passage sources directly from documents, it inherently ensures the validity of the information. Then we prompt LLMs along with the retrieved context Node 1 and 4 for the answer.

Additionally, for questions asking about document structures, we extract the document structure names and locate their corresponding structural nodes in the KG. For the table node, we retrieve its markdown formatted content while for the page node, we traverse its one-hop neighbor and obtain passages belonging to that page.

\SetAlCapSty{}
\SetAlgoNoEnd
\begin{algorithm}[tb]
 \DontPrintSemicolon
 \small
 \KwIn{A question $q$ over a set of documents $\mathcal{D}$, the constructed KG $G = \{\mathcal{V}, \mathcal{E}, \mathcal{X}\}$ over $\mathcal{D}$, the fine-tuned LLM-guided graph traversal $f_{\text{GT}}$, the preset context budget $K$, the TF-IDF search function $g$.}
 
Initialize seed passages $\mathcal{V}^s = g(\mathcal{V}, \mathcal{X}, q)$

Initialize the retrieved passage queue $\mathcal{P} = [\{v_i\}|v_i \in \mathcal{V}^{s}]$
 
Initialize the candidate neighbor queue $\mathcal{C} = [\mathcal{N}_i|v_i \in \mathcal{V}^s]$

Initialize the retrieved passage counter $k = \sum_{\mathcal{P}_i \in \mathcal{P}}|\mathcal{P}_i|$
 
\While{queue $\mathcal{P}$ and queue $\mathcal{C}$ are not empty}{
        $\mathcal{P}_i \leftarrow \mathcal{P}.\text{dequeue}()$, $\mathcal{C}_i \leftarrow \mathcal{C}.\text{dequeue}()$

        $\mathcal{V}_i' = \text{Graph Traversal}(\{q\}\cup \mathcal{P}_i, \mathcal{C}_i, k)$ by Eq~\eqref{eq-lmtraversal} 

        \For{$v \in \mathcal{V}_i'$}{
            $\mathcal{P}.\text{enqueue}(\mathcal{P}_i\cup \{v\})$, $\mathcal{C}.\text{enqueue}(\mathcal{N}_v)$

            $k \leftarrow k + 1$

            \If{$k > K$}{
                \textbf{Terminate}
            }
        }
    }
\KwRet{\text{Retrieved Passage Queue $\mathcal{P}$}}

\caption{\small LLM-based KG Traversal Algorithm to Retrieve Relevant Context for Content-based Question.}
\label{alg-GPL}
\end{algorithm}

Here we present the algorithm for our proposed KGP method for MD-QA. Given a question, we first apply LLM to classify whether the question is asking about the document structure or content. If the question focuses on the document structure, we extract the structural keywords such as Page or Table, and retrieve the content in the corresponding structural nodes in KG. If the question focuses on the document content, we follow the step according to Algorithm~\ref{alg-GPL}. Specifically, we first initialize seeding passages $\mathcal{V}^s$ and the reasoning path queue $\mathcal{P}$ by TF-IDF search. Then for each seeding passage $v_i \in \mathcal{V}^s$, we add its neighboring passage nodes $\mathcal{N}_i$ into the candidate neighbor queue $\mathcal{C}$ (lines 1-4).  After that, we iteratively dequeue the earliest enqueued reasoning path/candidate neighborhood $\mathcal{P}_i/\mathcal{C}_i$ from $\mathcal{P}/\mathcal{C}$ and employ the fine-tuned LLM-based graph traversal agent to rank the dequeued neighbors in $\mathcal{C}_i$ by Eq.~\eqref{eq-lmtraversal} (lines 5-7). Last, we select top-k passage nodes $\mathcal{V}_i'$ from $\mathcal{C}_i$ to visit next based on their rank and correspondingly update the candidate neighbor queue and reasoning path queue (lines 8-13). The above process terminates when either the candidate neighbor queue becomes empty or the prefixed budget $K$ for the retrieved passages is met. The time and space complexity are thoroughly analyzed in Section~\ref{sup-KGP} in Supplementary.

%% file: Experiment.tex
\begin{table*}[t!]
\centering
\scriptsize
\caption{MD-QA Performance (\%) of different baselines. The best and runner-up are in \textbf{bold} and \underline{underlined}. None: no passages but only the question is provided. Golden: supporting facts are provided along with the question.}
\begin{tabular}{l|ccc|ccc|ccc|ccc|c|cc}
\Xhline{2\arrayrulewidth}
\multirow{2}{*}{\textbf{Method}} & \multicolumn{3}{c|}{\textbf{HotpotQA}} & \multicolumn{3}{c|}{\textbf{IIRC}} & \multicolumn{3}{c|}{\textbf{2WikiMQA}} & \multicolumn{3}{c|}{\textbf{MuSiQue}} & \textbf{PDFTriage} & \multicolumn{2}{c}{\textbf{Rank}} \\

 & Acc & EM & F1 & Acc & EM & F1 & Acc & EM & F1 & Acc & EM & F1 & Struct-EM & w PDFTriage & w/o PDFTriage\\
\Xhline{2\arrayrulewidth}
None & 41.80 & 19.00 & 30.50 & 19.50 & 8.60 & 13.17 & 44.40 & 18.60 & 25.07 & 30.40 & 4.60 & 10.58 & 0.00 & 8.53 & 9.00\\
\hline
KNN & 71.57 & 40.73 & 57.97 & 43.82 & 25.15 & 37.24 & 52.40 & 31.20 & 42.13 & 44.70 & 18.86 & 30.04 & -- & 7.00 & 7.33\\
TF-IDF & \textbf{76.64} & \underline{45.97} & 64.64 & 47.47 & 27.22 & 40.80 & 58.40 & 34.60 & 44.50 & 44.40 & 21.59 & 32.50 & -- & 4.85 & 5.00\\
BM25 & 71.95 & 41.46 & 59.73 & 41.93 & 23.48 & 35.55 & 55.80 & 30.80 & 40.55 & 44.47 & 21.11 & 31.15 & -- & 6.92 & 7.25\\
\hline
DPR & 73.43 & 43.61 & 62.11 & 48.11 & 26.89 & \underline{41.85} & 62.40 & 35.60 & 51.10 & 44.27 & 20.32 & 31.64  & -- & 5.31 & 5.50\\
MDR & 75.30 & 45.55 & \underline{65.16} & \textbf{50.84} & \underline{27.52} & \textbf{43.47} & \underline{63.00} & 36.00 & \underline{52.44} & \underline{48.39} & \underline{23.49} & \underline{37.03}  & -- & \underline{3.07} & \underline{3.08}\\
\hline
IRCoT & 74.36 & 45.29 & 64.12 & \underline{49.78} & \textbf{27.73} & 41.65 & 61.81 & \underline{37.75} & 50.17 & 45.14 & 22.46 & 34.21 & -- & 4.00 & 4.08\\
\hline
KGP-T5  & \underline{76.53} & \textbf{46.51} & \textbf{66.77} & 48.28 & 26.94 & 41.54 & \textbf{63.50} & \textbf{39.80} & \textbf{53.50} & \textbf{50.92} & \textbf{27.90} & \textbf{41.19} & \textbf{67.00} & \textbf{2.69} & \textbf{2.75}\\
\hline
Golden & 82.19 & 50.20 & 71.06 & 62.68 & 35.64 & 54.76 & 72.60 & 40.20 & 59.69 & 57.00 & 30.60 & 47.75 & 100.00 & 1.00 & 1.00\\
\Xhline{2\arrayrulewidth}
\end{tabular}
\label{tab-mainperform10}
\vspace{-4ex}
\end{table*}

\section{Experiment}\label{sec-experiment}
In this section, we conduct experiments to verify the proposed knowledge graph prompting method (KGP) for MD-QA. In particular, we answer the following questions:
\begin{itemize}
\item \textbf{Q1 - Section~\ref{sec-performance}}: How well does KGP perform MD-QA compared with existing baselines?

\item \textbf{Q2 - Section~\ref{sec-graphimpact}-\ref{sec-LLMimpact}}: How do the quality of the constructed KG and the LLM-based graph traversal agent impact the MD-QA performance?
\end{itemize}

\noindent Due to the space limitation, we comprehensively introduce our experimental setting, including dataset collection, baselines, and evaluation criteria, in Supplementary~\ref{sup-datacollect}-\ref{sup-exprdetail}.

\begin{figure}[t!]
    \centering
    \includegraphics[width=0.48\textwidth]{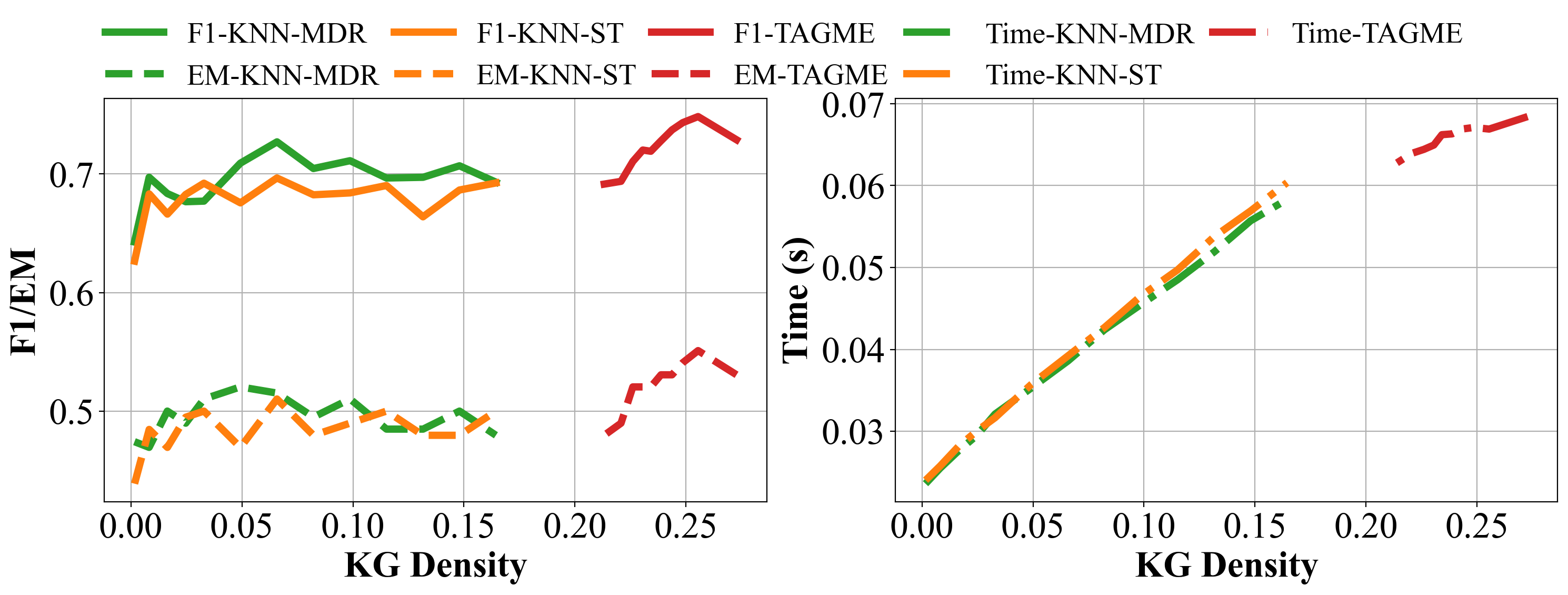}
    
    \caption{The performance/latency increases as the KG density increases. The results are averaged across 100 randomly sampled questions on HotpotQA.}
    \label{fig-hotpotqa_graph_perform}
    \vspace{-3ex}
\end{figure}

\subsection{Performance Comparison on MD-QA}\label{sec-performance}
We compare the MD-QA performance of the proposed KGP-T5 and other baselines in Table~\ref{tab-mainperform10}. Firstly, the baselines `None/Golden' achieve the worst/best performance because one provides no context and the other provides the golden context. All other baselines achieve the performance in-between because the retrieved context only covers the partial of the golden supporting facts. Our proposed methods KGP-T5 rank at the Top-1 except for the Golden baseline. The 2$^\text{nd}$-performing baseline MDR fine-tunes a RoBERTa-base encoder by predicting the next supporting fact based on the question and the already retrieved contexts~\cite{xiong2020answering}. This next-passage prediction pretext task equips the model with the reasoning capability of the knowledge across different passages and hence increases the quality of the retrieved contexts. The other deep-learning-based retriever DPR achieves much worse performance than MDR because it only fine-tunes the encoder by maximizing the similarity between the query and its supporting facts regardless of their sequential order, demonstrating the importance of understanding the logical order of different knowledge when solving MD-QA~\cite{xiong2020answering}. By comparing the MD-QA performance across different datasets, we find that all baselines perform better on HotpotQA than on IIRC. This is because questions in HotpotQA are generally simpler than in IIRC. Existing works~\cite{jiang2019avoiding} have shown that some questions in HotpotQA can be easily answered following shortcuts while questions in IIRC sometimes necessitate arithmetic skills to derive answers numerically, e.g., `How many years did the event last when Wingfield lost his fortune?', which poses unique difficulty due to LLMs' inferior arithmetic capability~\cite{yuan2023well}.

Moreover, without any particular design for document structures, no existing baselines can handle structural questions in PDFTriage, e.g. `What is the difference between Page 1 and Page 2' or `In Table 3, which station has the highest average flow rate?'. Fortunately, with the constructed KG incorporating structural nodes and our designed traversal algorithm retrieving structural contexts, our proposed method achieves 67\% Struct-EM.


\begin{figure*}[t!]
    \centering
    \includegraphics[width=1\textwidth]{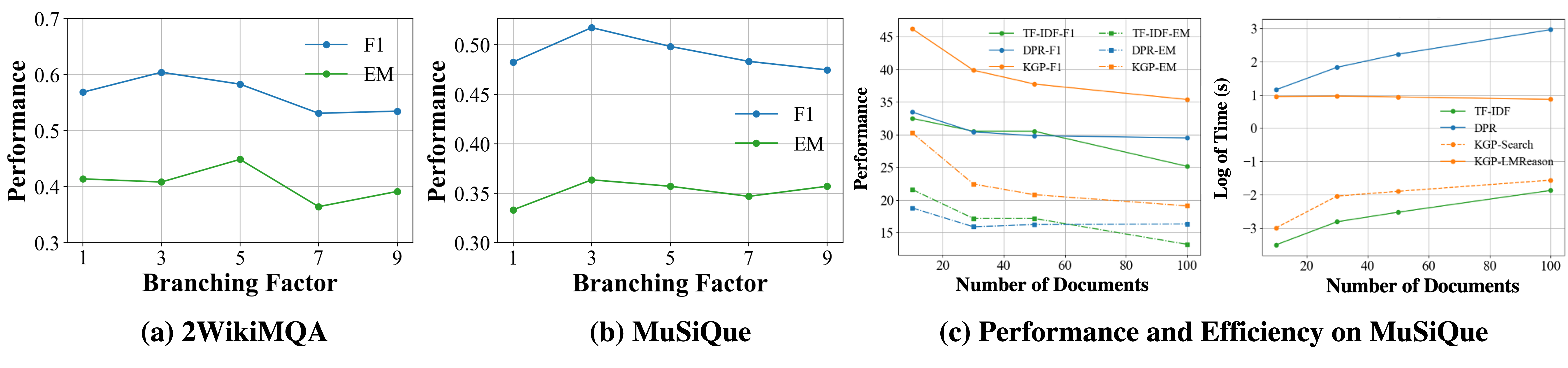}
    \vspace{-5ex}
    \caption{\textbf{(a)-(b)}: Performance first increases and then decreases as the branching factor increases. The results are averaged across 100 sampled questions on 2WikiMQA and MuSiQue. \textbf{(c)}: Performance/Efficiency increases/decreases as the number of documents increases on MuSiQue. KGP-T5 achieves higher performance/efficiency than DPR.}
    \vspace{-2ex}
    \label{fig-analysis}
\end{figure*}

\begin{table}[t!]
\vspace{-2ex}
\tiny
\setlength\tabcolsep{2pt}
\caption{Comparing different LLM-based KG Traversal Agents, including off-the-shelf ChatGPT equipped with few-shot demonstration with fine-tuned LLaMA/T5/MDR on TAGME-constructed KG.}
\begin{tabular}{l|ccc|ccc|ccc|ccc}
\hline
 \textbf{Traversal} & \multicolumn{3}{c|}{\textbf{HotpotQA}} & \multicolumn{3}{c|}{\textbf{IIRC}} & \multicolumn{3}{c|}{\textbf{2WikiMQA}} & \multicolumn{3}{c}{\textbf{MuSiQue}} \\
 \textbf{Agent}  & Acc & EM & F1 & Acc & EM & F1 & Acc & EM & F1 & Acc & EM & F1 \\
\hline
TF-IDF & 73.52 & 43.79 & 63.14  & 46.30  & 27.70  & 41.43  & 58.12  & 35.07 & 45.95 & 44.67 & 21.93 & 32.90\\
\hline
MDR & 75.72 & 46.09 & 65.77 & \textbf{49.58} & \textbf{29.32} & \textbf{43.21} & 60.94 & 37.22 & 51.29 &  \textbf{51.22} & \underline{27.76} & \underline{41.11}\\
\hline
ChatGPT & \textbf{77.80} & 46.03 & \underline{66.57} & 46.27 & 26.01 & 39.35 & 61.62 & 36.16 & 49.39 & 50.61 & 26.92 & 38.66 \\
LLaMA & 75.66 & \underline{46.22} & 66.31 & \underline{49.57} & \underline{28.09} & \underline{42.56} & \underline{62.45} & \underline{37.55} & \underline{52.45} & 50.81 & 26.72 & 40.01 \\
T5 & \underline{76.53} & \textbf{46.51} & \textbf{66.77} & 48.28 & 26.94 & 41.54 & \textbf{63.50} & \textbf{39.80} & \textbf{53.50} & \underline{50.92} & \textbf{27.90} & \textbf{41.19} \\
\hline
\end{tabular}
\label{tab-ablation}
\vspace{-4ex}
\end{table}

\subsection{Impact of the Constructed Graph}\label{sec-graphimpact}
We construct KGs with varying densities by varying the hyperparameters of TF-IDF/KNN-ST/KNN-MDR/TAGME, and studying its impact on the performance and the neighbor matching time of MD-QA using KGP-T5. Since the LLM-based graph traversal agent selects the next node to visit from neighbors of already visited nodes, the chance that it hits the supporting facts increases as neighbors increase. In contrast, the neighbor matching efficiency decreases as the candidate pool, i.e., $\mathcal{N}_j$ in Eq~\eqref{eq-lmtraversal}, increases. As evidenced in Figure~\ref{fig-hotpotqa_graph_perform}, we observe a similar trend, i.e., as KG density increases, the F1/EM increases and stays stable while the latency for selecting the most promising neighbors to visit next also increases. KNN-MDR achieves better performance than KNN-ST when the density of the two constructed KGs is the same. This is because the encoder in KNN-ST is pre-trained on wide-spectrum datasets while the encoder in MDR is specifically pre-trained on the HotpotQA by the pretext task of predicting the next supporting facts. Therefore, the embedding similarity and the corresponding neighbor relations better reflect the logical associations among different passages, which aligns with the better constructed KG by KNN-MDR than KG by KNN-ST in Figure~\ref{fig-hotpotqa_graph}. Compared with KNN-MDR/ST, TAGME delivers superior performance at the cost of increasing latency since the generated KG by TAGME is denser than KGs by KNN-ST/MDR.

\vspace{-1ex}
\subsection{Impact of Graph Traversal Agent}\label{sec-LLMimpact}
Here we study the influence of using different LLM agents to traverse over TAGME-constructed KG on MD-QA. Specifically, we compare agents that select the next neighbor to visit randomly or intelligently via guidance from ChatGPT, LLaMA, T5, and MDR in Table~\ref{tab-ablation}. Because the random agent only blindly traverses the KG without any guidance from LLM, it unavoidably collects irrelevant passages and hence achieves the worst performance than others under LLMs' guidance. This aligns with our previous observation on the low precision in Figure~\ref{fig-hotpotqa_graph} and further demonstrates the necessity of using LLMs to guide the graph traversal. Interestingly, we find that KGP-T5 performs better than LLaMA even though the parameters of LLaMA-7B are more than the ones with T5-0.7B. We hypothesize this is because LLaMA-7B requires more data to fine-tune than T5-0.7B.

\vspace{-1ex}
\subsection{Sensitivity Analysis}\label{sec-sensi}
Here we perform the sensitivity analysis of the branching factor (the number of nodes selected from candidate neighbors to visit next). In Figure~\ref{fig-analysis}(a)-(b), the performance first increases as the branching factor increases because more passage nodes selected from the candidate neighbors lead to more reasoning paths to reach the final answer. However, as we fix the context budget to ensure fair comparison (i.e., the total number of passages we are allowed to retrieve for each question is the same across all baselines), the performance declines as the branching factor increases because the number of initial seeding nodes diminishes, leading to reduced coverage of the KG. Furthermore, we compare the efficiency of KGP when the constructed KG includes different numbers of documents in Figure~\ref{fig-analysis}(c). KGP consistently achieves higher performance than other baselines and higher efficiency than embedding-based DPR. TF-IDF is slightly faster than KGP because it is a purely heuristic-based method.

%% file: related.tex
\section{Related Work}
\textbf{Question answering} 
Question Answering (QA) aims to provide answers to users' questions in natural language~\cite{zhu2021retrieving, pandya2021question}, and most QA systems are composed of information retrieval (IR) and answer extraction (AE)~\cite{mao2021rider, ju2022grape, liu2022heterogeneous}. In IR, the system searches for query-relevant factual passages using heuristic methods (BM25)~\cite{robertson2009probabilistic} or neural-ranking ones (DPR)~\cite{karpukhin2020dense}. In AE, the final answer is extracted usually as a textual span from related passages. Although this framework has been broadly applied in O-QA~\cite{mao2021rider} and D-QA~\cite{xu2020layoutlmv2, mathew2021docvqa}, no previous work focus on MD-QA, which demands alternatively reasoning and retrieving knowledge from multiple documents. To tackle this issue, we construct KGs to encode logical associations among different passages across documents and design an LLM-based graph traversal agent to alternatively generate the reason and visit the most matching passage node.

\noindent\textbf{Pre-train, Prompt, and Predict with LLMs}
With the emergence of LLMs, the paradigm of `pre-train, prompt, predict' has gained magnificent popularity in handling a wide spectrum of tasks~\cite{gururangan2020don, liu2023pre, yu2023large}. This approach begins with pre-training LLMs by pretext tasks to encode world knowledge into tremendous parameters~\cite{wu2023survey} followed by a prompting function to extract pertinent knowledge for downstream tasks~\cite{yang2023harnessing}. Recent advancements explore different prompting strategies to enhance LLMs' reasoning capabilities~\cite{wei2022chain, jin2023large}. In contrast to that, our work offers a novel perspective by transforming the prompt formulation into the KG traversal. 

%% file: conclusion.tex
\section{Conclusion}
Answering multi-document questions demands knowledge reasoning and retrieving from different documents across various modalities, presenting challenges for applying the paradigm of `pre-train, prompt and predict' with LLMs. Recognizing the logical associations among passages and structural relations within documents, we propose a Knowledge Graph Prompting method (KGP) for aiding LLMs in MD-QA. The KGP constructs KGs from documents with nodes as sentences or document structures, and edges as their lexical/semantic similarity/structural relations. Since constructed KGs may contain irrelevant neighbor information, we further design an LLM-based graph traversal agent that selectively visits the most promising node in approaching the question. In the future, we plan to investigate the capability of LLMs in understanding graph topology and explore the potential of fine-tuning/prompting LLMs to encode complex topological signals hidden in the graph.

%% file: supplement.tex
\clearpage
\newpage
\section{Supplementary}
\subsection{Dataset Collection}\label{sup-datacollect}
This section introduces the collection of datasets used for the experiments conducted in this paper.

\subsubsection{Document Set Collection and Procession}
As no previous works focus on MD-QA, we create our own datasets to simulate real-world scenarios where users maintain folders containing various documents and pose questions to which the answers are only from certain parts of these documents. To imitate this scenario, we randomly sample questions from the development set of existing datasets: HotpotQA/IIRC/2WikiMQA/MuSiQue, and then for each specific question, we fetch documents from Wikipedia that encompass supporting facts pertaining to the question~\footnote{The HotpotQA/IIRC/2WikiMQA/Musique datasets already have the supporting facts for each question.} and term these documents as golden documents. Then we randomly sample negative documents from Wikipedia and pair them with golden documents to constitute the document collection. For each document in the collected document set, we split it into multiple passages with the default passage length being 250 as it empirically yields superior performance. As questions from these existing datasets only focus on document contents, we additionally incorporate the `PDFTriage' dataset, an internal company collection of real-world questions focusing on document structures. We refer readers to the paper~\cite{saad2023pdftriage} for more details.

\subsubsection{Knowledge Graph Construction}
We construct a knowledge graph for each question and its corresponding collection of documents. For datasets where the questions are from Wikipedia: HotpotQA, IIRC, WikiMHop, and Musique, we only have passage nodes since answering questions in these datasets does not require information about document structures. For the PDFTriage dataset, in addition to passage nodes, we apply ExtractAPI to obtain the page and table information so that the constructed KG also has pages/tables as nodes. For all of these datasets, we add edges following Section~\ref{sec-kgconstruct}. Table~\ref{tab-dataset} summarizes the average statistics of the document collections across all questions with their corresponding KGs. The code for the dataset collection and preprocessing is publically available at \url{https://github.com/YuWVandy/KG-LLM-MDQA}.

\begin{table}[htbp!]
\vspace{-2ex}
\scriptsize
\setlength\tabcolsep{2pt}
\caption{Statistics of document collections and their corresponding knowledge graph used in Table~\ref{tab-mainperform10} and \ref{tab-ablation} average across all questions.}
\centering
\begin{center}
\begin{tabular}{lcccccc}
\Xhline{2\arrayrulewidth}
\textbf{Dataset} & \#\textbf{Docs} & \#\textbf{Questions} & \#\textbf{Passages} & \#\textbf{Edges}& \textbf{\makecell{Passage \\Avg. Length}} & \textbf{\makecell{KG \\Density}} \\
\Xhline{2\arrayrulewidth}
HotpotQA & 12 & 500 & 715.22 & 70420.68 & 37.55 & 0.23\\
IIRC & 12 & 477 & 1120.55 & 143136.17 & 37.24 & 0.20\\
WikiMHop & 12  & 500 & 294.19 & 19235.15 & 37.24 & 0.27 \\
MuSiQue & 12 & 500 & 748.04 & 97931.28 & 38.56 & 0.29 \\
\Xhline{2\arrayrulewidth}
\end{tabular}
\end{center}
\label{tab-dataset}
\vspace{-2ex}
\end{table}

More details about the PDFTriage dataset can be found at PDFTriage~\cite{saad2023pdftriage}.

\subsubsection{Sequential Data Collection} Training MDR~\cite{xiong2020answering} requires rearranging supporting facts into the sequential order that progressively approaches the answer. To fulfill this requirement, we directly follow MDR and use the pre-processed HotpotQA data from the GitHub Repository\footnote{https://github.com/facebookresearch/multihop\_dense\_retrieval/tree/main} to train the encoder and apply it to other datasets that do not provide the sequential order of supporting facts. For instruction fine-tuning LLaMA, we still use the above HotpotQA data and rearrange it into the instruction-input-output format and use the instruction `What evidence do we need to answer the question given the current evidence'. We present one example in Listing~\ref{list-InstructData}. For T5-large, we use the same input-output but prefix the reasoning instruction to the input following the original T5 input format~\cite{raffel2020exploring}.

\subsection{Experiment Details}\label{sup-exprdetail}
\subsubsection{Training DPR and MDR}
For training DPR~\cite{karpukhin2020dense}, we pair each question with its supporting facts as its positive passages, and some randomly sampled passages as its negative passages. For training MDR~\cite{xiong2020answering}, as each question in HotpotQA only requires 2 supporting facts to derive the answer, we set the first supporting fact as the positive pair for each question. Further, we concatenate this question and the first supporting fact to form a new question and for this newly-formed question, we set the second supporting fact as its positive pair. For both the original question and the concatenated one, we randomly sample other passages as the negative pair. Following~\cite{xiong2020answering, karpukhin2020dense}, we use RoBERTa-base as the default encoder. The search space of hyperparameters is summarized in Table~\ref{sup-tab-hyper-mdrdpr}.

\begin{table}[htbp!] 
\vspace{-1ex}
\footnotesize
\setlength{\extrarowheight}{0.5pt}
\setlength\tabcolsep{3pt}
\caption{Hyperparameters used for tuning DPR and MDR. The value of most of them are directly taken from their original GitHub Repository.}
\vspace{-2ex}
\label{sup-tab-hyper-mdrdpr}
\begin{center}
\begin{tabular}{m{13em} m{13em}}
\Xhline{2\arrayrulewidth}
\textbf{Hyperparameter} &  \textbf{Search Space}\\
\Xhline{2\arrayrulewidth}
Encoder & RoBERTa-base\\
Hidden Dimension & 768\\
Max Context Length & \{128, 256, 350\}\\
Batch Size & \{128, 256, 512\}\\
Epoch & 50\\
Warmup Steps & 300 \\
Learning Rate & 2e-5 \\
Gradient Clipping Range & 2 \\
\Xhline{2\arrayrulewidth}
\end{tabular}
\end{center}
\end{table}

\subsubsection{Instruction Fine-tuning LLaMA\footnote{\url{https://github.com/Lightning-AI/lit-llama}} and T5-Large\footnote{\url{https://shivanandroy.com/fine-tune-t5-transformer-with-pytorch/}}} We fine-tune LLaMA using instruction data in Listing~\ref{list-InstructData}. Due to the computational limitation, we choose LLaMA-7B and use LoRA~\cite{hu2021lora}. For fine-tuning T5-Large, we use the same instruction data except that we remove the instruction but only prefix the reasoning instruction to the input~\cite{raffel2020exploring}. We use the default hyperparameters from their original GitHub repository to fine-tune these two LLMs.

\subsubsection{Prompting LLMs for MD-QA - Table~\ref{tab-mainperform10} and \ref{tab-ablation}}
Following~\cite{trivedi2022interleaving}, we randomly select questions from the development set for reporting the performance. To ensure a fair comparison, we set the number of retrieved passages to 30 across all baselines and use ChatGPT as the downstream LLM for reading the retrieved passages and generating the answer. We summarize the key implementation details for each baseline as follows:
\begin{itemize}[leftmargin=*]
    \item \textbf{KNN}: We employ the sentence-transformer variant `multi-qa-MiniLM-L6-cos-v1' to obtain passage embeddings as it has been trained on 215M (question, answer) pairs from diverse sources. Then we select the top-15 passages according to the embedding similarity and the top-15 passages according to the fuzzy matching\footnote{We use Levenshtein-distance to measure the lexical distance between two passages.}.

    \item \textbf{MDR}: We use beam search with the inner product as the scoring function to rank passages. We limit the search depth to 2 as answering questions in HotpotQA requires at most 2-hop reasoning steps~\cite{xiong2020answering}. We set the number of passages to be 15 in the first-hop retrieval and for each of these passages, we further retrieve 3 more passages in the second round, which in total generates 45 passage pairs. Then we rank these 45 passage pairs by the product of the scores between the first-hop and the second-hop retrieval and select the top 30 ones as the final context.

    \item \textbf{IRCoT}: Instead of directly employing the original IRCoT code~\cite{trivedi2022interleaving}, we modify it based on our problem setting. The first reason is that passages to be retrieved in IRCoT~\cite{trivedi2022interleaving} are the pre-processed Wikipedia Corpus and do not cover the whole contents of Wikipedia documents, which thereby is not aligned with our MD-QA setting. The second reason is that the question-answering reader employed in IRCoT requires running on A100-80G GPU, which is not affordable on our side. Therefore, we modify the IRCoT by replacing the question reader with the ChatGPT and using our pre-processed Wikipedia document collections as introduced in Section~\ref{sup-datacollect}. For the prompt used in the reasoning step, we select 2 examples from `gold\_with\_2\_distractors\_context' for the demonstration purpose. We iteratively select top-5 passages based on the generated reason from LLM along with their document titles and add them to the retrieved context until hitting the prefix budget. For the prompt used in the reading step, we use exactly the same prompt as other baselines as we find it empirically leads to better performance than the original one used in IRCoT~\cite{trivedi2022interleaving}.

    \item \textbf{KGP-T5/LLaMA/MDR/ChatGPT}: We use T5-large/LLaMA-7B/MDR/ChatGPT as the LLM to guide the graph traversal respectively. For content-based questions, similar to MDR, we perform a 2-hop retrieval but for each hop, we only search the node to visit next from neighbor candidates. In the 1$^{\text{st}}$-hop retrieval, we select 10 passages and in 2$^{\text{nd}}$-hop retrieval, we select 3 passages, which totally forms 30 reasoning paths. Note that passages in the 1$^{\text{st}}$-hop retrieval are allowed to overlap with the ones in the 2$^{\text{nd}}$-hop retrieval. For structural-based questions, we first use ChatGPT to extract page/table structures and then fetch relevant contents in those structures. Future work could explore how to pre-train a structural extraction model to obtain document structures.

    \item \textbf{KGP-TF-IDF}: We remove the LLM-guided graph traversal but select passage nodes based on their TF-IDF similarity to the given question.

\end{itemize}

Note that we put the prompt template for running all the above baselines in Section~\ref{sec-prompt}.

\subsection{Complexity Analysis for KGP}\label{sup-KGP}

\SetAlCapSty{}
\SetAlgoNoEnd
\begin{algorithm}[htbp!]
 \DontPrintSemicolon
 \small
 \KwIn{A question $q$ over a set of documents $\mathcal{D}$, the constructed KG $G = \{\mathcal{V}, \mathcal{E}, \mathcal{X}\}$ over $\mathcal{D}$, the fine-tuned LLM-guided graph traversal $f_{\text{GT}}$, the preset context budget $K$, the TF-IDF search function $g$.}
 
Initialize seed passages $\mathcal{V}^s = g(\mathcal{V}, \mathcal{X}, q)$

Initialize the retrieved passage queue $\mathcal{P} = [\{v_i\}|v_i \in \mathcal{V}^{s}]$
 
Initialize the candidate neighbor queue $\mathcal{C} = [\mathcal{N}_i|v_i \in \mathcal{V}^s]$

Initialize the retrieved passage counter $k = \sum_{\mathcal{P}_i \in \mathcal{P}}|\mathcal{P}_i|$
 
\While{queue $\mathcal{P}$ and queue $\mathcal{C}$ are not empty}{
        $\mathcal{P}_i \leftarrow \mathcal{P}.\text{dequeue}()$, $\mathcal{C}_i \leftarrow \mathcal{C}.\text{dequeue}()$

        $\mathcal{V}_i' = \text{Graph Traversal}(\{q\}\cup \mathcal{P}_i, \mathcal{C}_i, k)$ by Eq~\eqref{eq-lmtraversal} 

        \For{$v \in \mathcal{V}_i'$}{
            $\mathcal{P}.\text{enqueue}(\mathcal{P}_i\cup \{v\})$, $\mathcal{C}.\text{enqueue}(\mathcal{N}_v)$

            $k \leftarrow k + 1$

            \If{$k > K$}{
                \textbf{Terminate}
            }
        }
    }
\KwRet{\text{Retrieved Passage Queue $\mathcal{P}$}}

\caption{\small LLM-based KG Traversal Algorithm to Retrieve Relevant Context for Content-based Question.}
\end{algorithm}

Since our algorithm can be essentially deemed as the combination of the neighborhood ranking by Eq.~\eqref{eq-lmtraversal} and the breadth-first-search. The time complexity would be the multiplication between the time of bread-first-search $\mathcal{O}(|\mathcal{V}| + |\mathcal{E}|)$ and the time of neighborhood ranking $\mathcal{O}(|\mathcal{N}|\gamma) = \mathcal{O}(\hat{d}\gamma)$ where $\gamma$ is the time for computing the embedding similarity between a specific neighbor passage and the retrieved reasoning path and $\hat{d}$ is the average degree of the KG. Therefore the final time complexity would be $\mathcal{O}((|\mathcal{V}| + |\mathcal{E}|)\hat{d}\gamma)$, which is in-between the linear and quadratic to the size of the graph. As users typically maintain 10-100 documents, correspondingly the number of nodes in the constructed KG would be around 1,000-10,000 (according to Table~\ref{tab-dataset}, a collection of 12 documents have roughly 200-1000 passage nodes), which is affordable even with the quadratic time complexity. Moreover, we can apply advanced techniques to further reduce the time complexity for neighborhood ranking, such as LSH~\cite{gionis1999similarity} and KD-tree~\cite{qu2020data}.

In addition, whenever there are some changes over the document set (e.g., the user adds a new document into the folder or removes an existing document), we can remove/add all sentence nodes from/to the graph. To guarantee the linear time complexity for removing sentences from one document, we need to maintain a pointer from the document to its sentence nodes. For adding sentence nodes of one document, we need to first apply the KG construction method to compute the lexical/semantic similarity between each of the newly added sentence nodes and the existing nodes in KG, and then add corresponding edges connecting them, which is also linear to the size of the current graph.

For space complexity, it takes $\mathcal{O}(|\mathcal{V}|(\alpha + \beta))$ to maintain the constructed KG on the fly where $\alpha$ is the average space for saving the passage embedding vector while $\beta$ is the average space for saving the textual information of that passage. Although our constructed KG treats passages as nodes, which cannot scale very well when the graph is extremely large, the total number of documents a user maintains in a folder is typically around 10-100, which is still affordable.

\subsection{Markdown-Formatted Table}\label{sup-KGP-table}
\begin{figure}[htbp!]
    \centering
    \includegraphics[width=0.48\textwidth]{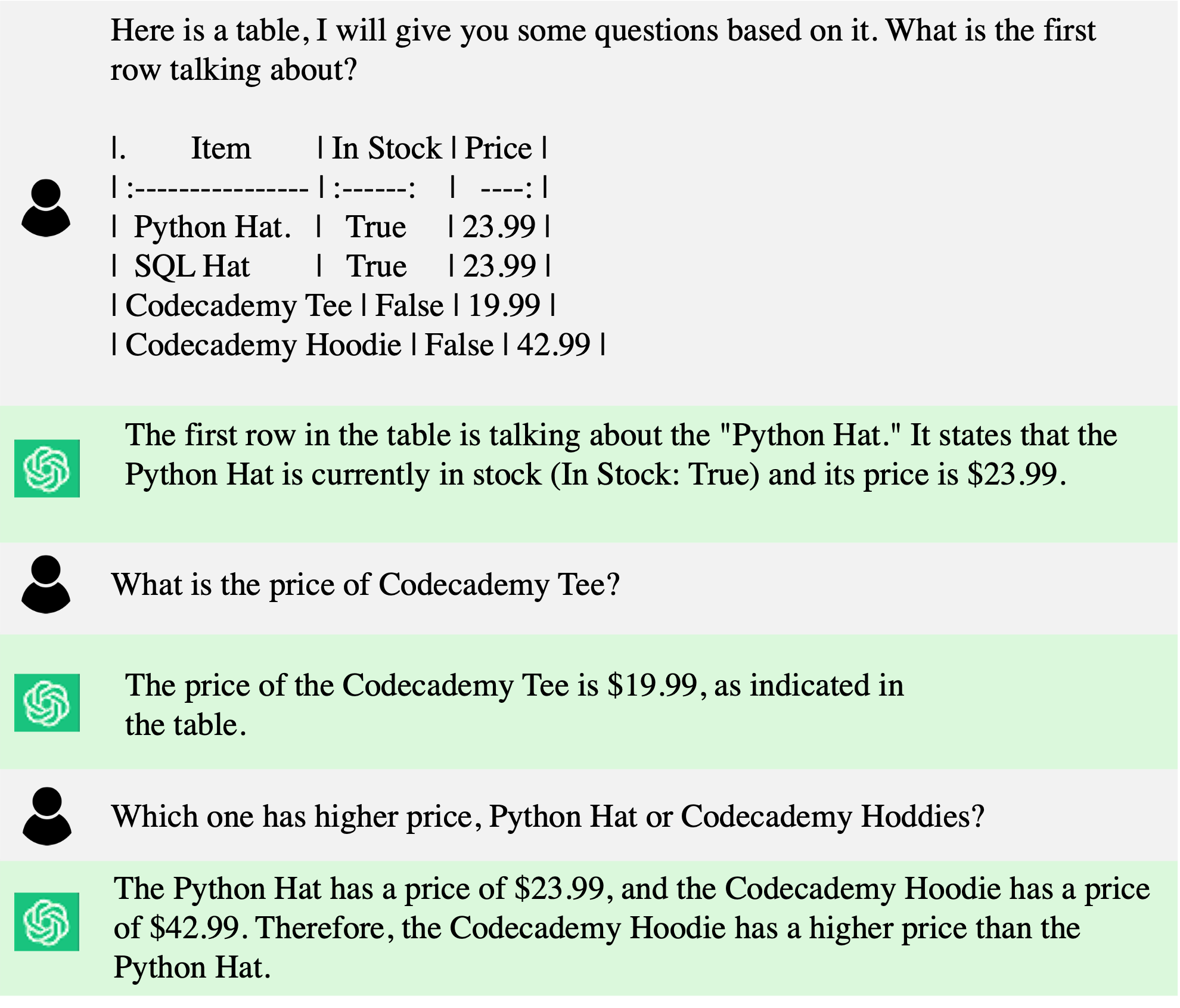}
    \caption{An example demonstrating that ChatGPT can understand table in the markdown format.}
    \label{fig-tableunderstanding}
\end{figure}

Figure~\ref{fig-tableunderstanding} demonstrates that by sending Tables in the markdown format, ChatGPT can successfully understand their content and perform information retrieval based on the given questions. However, we do observe that such a markdown-formatted solution is not feasible for the long table due to the input token limitation of ChatGPT, we plan to explore the solution using SQL as the prompt content or modeling the Table as the grid graph to solve the issue in the future.

\begin{table*}[t!]
\scriptsize
\setlength{\extrarowheight}{.095pt}
\setlength\tabcolsep{3pt}
\centering
\caption{Systematically Comparison among existing and our proposed Knowledge Graphs.}\label{tab-compare}
\begin{tabular}{lcccccc|c|c}
\hline
\textbf{KG} & \textbf{Node}& \textbf{Edge} & \textbf{Domain} & \textbf{Constructor} & \textbf{Scalability} & \textbf{Hyperparameters} & \textbf{Advantage} & \textbf{Disadvantage}\\
 \hline
\textbf{TAGME} & Passage & \makecell{Common \\Wikipedia Entity} & Wikipedia & / & No & Prior Threshold & \makecell{Effectively Identify \\Wikipedia Entities} & \makecell{Low efficiency for Entity Identification \\Narrow Domain Application} \\  
\hline
\textbf{TF-IDF} & Passage & Common Keyword & General & / & No & \# Keywords & No Domain Limitation & Common keywords irrelevant to question\\ 
\hline

\textbf{KNN-ST} & Passage & Semantic Similarity & General & \makecell{Sentence\\Transformer} & No & \# Neighbors & No Domain Limitation & Semantic Similarity irrelevant to question\\ 
\hline
\textbf{KNN-MDR} & Passage & Semantic Similarity & General & MDR & No & \# Neighbors & \makecell{Encoding the logical \\association for QA} & \makecell{Require logically ordered \\supporting facts to pre-train the model}\\  
\hline
\textbf{\makecell{Knowledge\\Base}} & Entity & Relationship & Specific & Human & Yes & / & \makecell{Powerful in encoding\\ factual information} & \makecell{Relation Extraction is non-trivial\\Domain Specific}\\  
\hline
\end{tabular}
\label{tab-kgconstruct}
\end{table*}

\subsection{Knowledge Graph Construction Comparison}\label{sup-KGC}
Table~\ref{tab-kgconstruct} compares different knowledge graph construction methods and their pros and cons. 

\begin{itemize}
    \item \textbf{TAGME}: TAGME~\cite{ferragina2010tagme} is very effective in extracting Wikipedia Entities from a passage despite the low efficiency. In our graph construction, it usually takes more than 8 hours to extract entities of all passages for even just 12 Wikipedia documents. Even after we apply parallel processing, it still takes more than 2 hours. In addition, it can only handle entities mentioned in the existing Wikipedia system and hence cannot generalize to documents from other domains. 

    \item \textbf{TF-IDF and KNN-ST}: Although there is no domain limitation, it is hard to guarantee the extracted keywords or the embedding semantic similarity can precisely encode the relationships that are desired for answering the given question between any two passages. We empirically find TF-IDF is more likely to extract meaningless keywords even after removing supporting verbs and articles.

    \item \textbf{KNN-MDR}: Since KNN-MDR pre-trains the sentence encoder by predicting the next supporting passage given already-retrieved passages, the embedding similarity between two passages is more likely to encode necessary logical associations required for MD-QA. However, the main bottleneck here is how to obtain the logically ordered supporting facts that can progressively reach the answer. Obtaining these sequential data is non-trivial and usually requires a large number of human resources for well-curated annotation.

    \item \textbf{Existing Knowledge Base}: One common approach in the literature is to use existing knowledge bases or extract subgraphs from them for specific tasks~\cite{yasunaga2022deep, dong2023hierarchy, yasunaga2021qa}. Because the factual information is characterized as a triplet consisting of two entity nodes and their relationship, it is very powerful in encoding factual information/commonsense knowledge and also avoids the scalability issue (since two different passages might share the same entity). Despite its potency and ease of use, constructing this type of KGs demands meticulously designed relation extractors, which is still deemed a challenging task in the literature. Recent research has explored using LLMs for relation extraction. However, with increasing document numbers, using non-open-sourced LLMs can become prohibitively expensive. A potential solution is fine-tuning an open-sourced LLM specifically for relation extraction. Detailed discussion on this is beyond the scope of this study and is thus omitted.
\end{itemize}

To put it in a nutshell, there's no one-size-fits-all method for KG construction. Our paper offers an in-depth analysis of the proposed KG construction methods alongside other existing ones. The best approach often depends on the specific use case. For broad domains containing general factual information, tools like 'TAGME' or 'Knowledge Base' might be apt. However, for more niche or sensitive areas, methods like TF-IDF/KNN-ST are more appropriate. In certain situations, gathering domain-specific data and pre-training encoders is the most effective way to build the KG.

\newpage
\subsection{Additional Results and Discussions}\label{sup-res}
\subsubsection{Quality of KG on MuSiQue}
Similar to the setting used for Figure~\ref{fig-hotpotqa_graph}, we change the hyperparameters to construct KGs for each question in MuSiQue with varying levels of sparsity and measure how much percentage of the supporting facts are covered by neighbors of the seeding passages that are initially retrieved by TF-IDF. The general trend is similar to the one in Figure~\ref{fig-hotpotqa_graph}, i.e., as the graph becomes denser, the precision decreases while the SF-EM increases. However, on MuSiQue, KNN-MDR achieves the worst trade-off between Precision and SF-EM compared with KNN-ST and TF-IDF. This is because our KNN-MDR is pre-trained on HotpotQA and due to the distribution shift from HotpotQA to MuSiQue, it is expected for the graph constructed with KNN-MDR to have less quality. Note that although here KNN-ST leads to a better KG than KNN-MDR, it does not mean the KNN baseline in Table~\ref{tab-mainperform10} should perform better than MDR because the baseline name only refers to the retrieval method while the name in this figure refers to the KG construction method.

\begin{figure}[htbp!]
    \centering
    \includegraphics[width=0.48\textwidth]{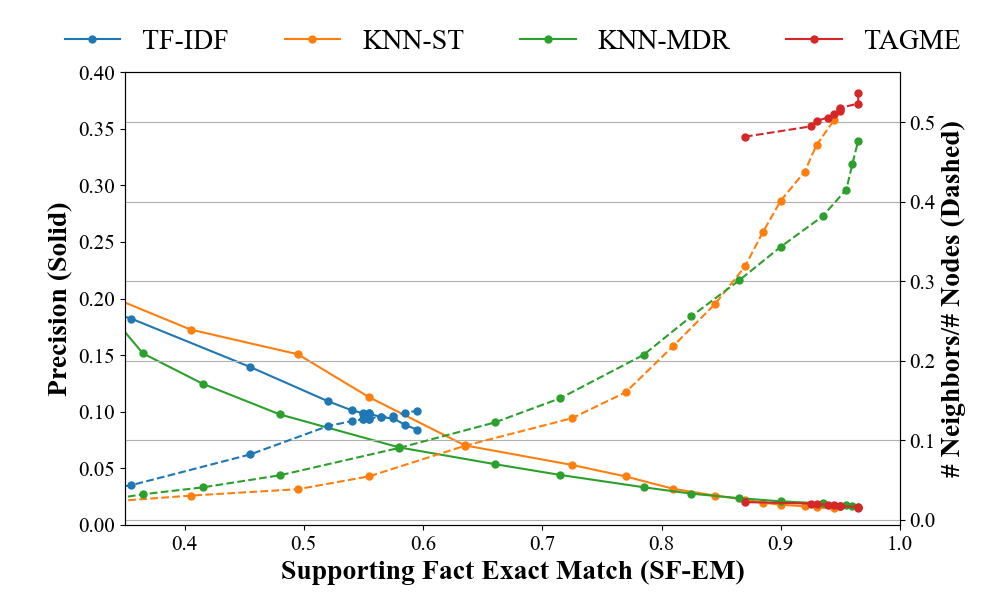}
    \caption{Quality of constructed KGs with different methods on MuSiQue. \textbf{TF-IDF}: lexical similarity based on common keywords extracted by TF-IDF. \textbf{KNN-ST}: KNN graph constructed based semantic similarity of embeddings from sentence-transformer; \textbf{KNN-MDR}: KNN graph constructed based on semantic similarity of embeddings from the pre-trained MDR~\cite{xiong2020answering}; \textbf{TAGME}: graph constructed based on whether two passages share common Wikipedia entity mentions}
    \label{fig-musique_graph}
\end{figure}

\subsubsection{The impact of KG on MuSiQue}
Similar to the setting used for Figure~\ref{fig-hotpotqa_graph_perform}, we compare the MD-QA performance for KGP-T5 using TAGME-based KG with different levels of density. Similar to Figure~\ref{fig-hotpotqa_graph_perform}, here we also observe that as the KG becomes denser, the MD-QA performance increases while the time for the next node search increases. However, on MuSiQue, in most cases, KNN-ST achieves better F1/EM than KNN-MDR, which exactly aligns with the constructed KG quality observed in Figure~\ref{fig-musique_graph}, i.e., KNN-ST achieves better Precision/SF-EM trade-off than KNN-MDR on MuSiQue.

\begin{figure}[t!]
    \centering
    \includegraphics[width=0.48\textwidth]{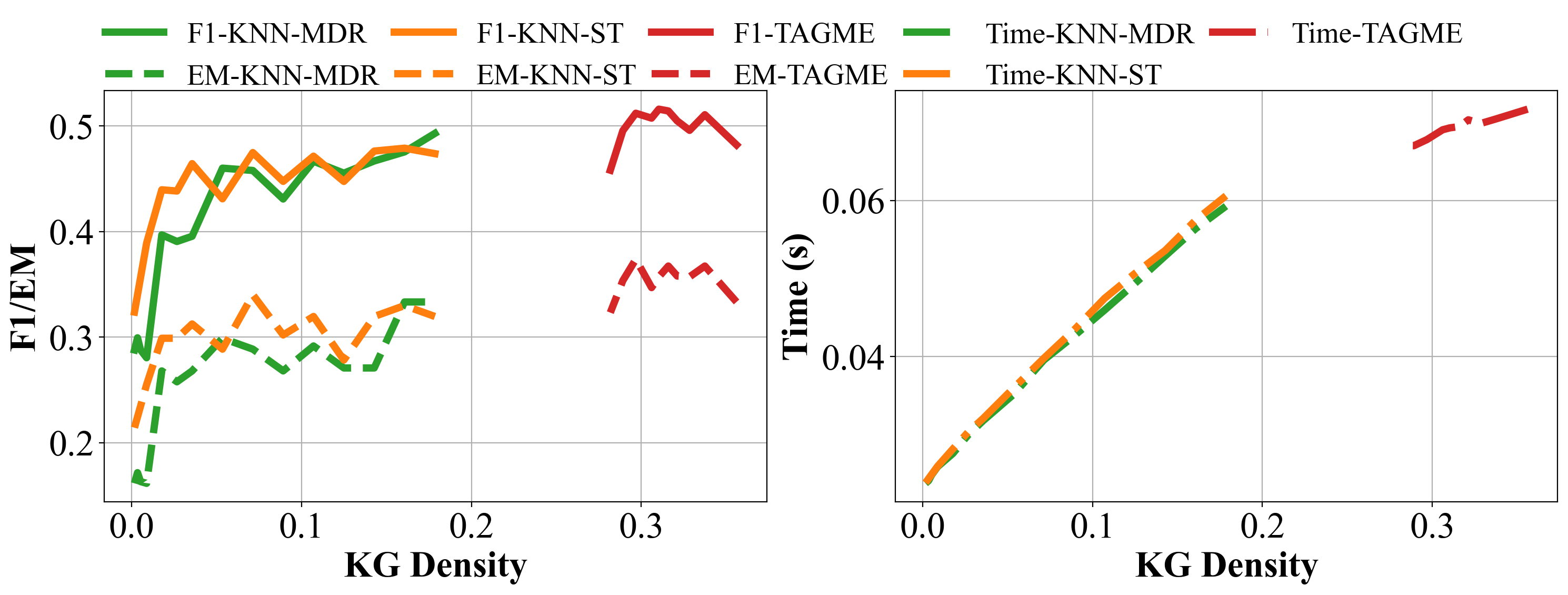}
    \caption{The performance/latency increases as the KG density increases. The results are averaged across 100 randomly sampled questions on MuSiQue.}
    \label{fig-musique_graph_perform}
\end{figure}

\subsection{Case study on Structural/Content Questions}\label{sec-casestudy}
\begin{figure}[htbp!]
    \centering
    \includegraphics[width=\textwidth]{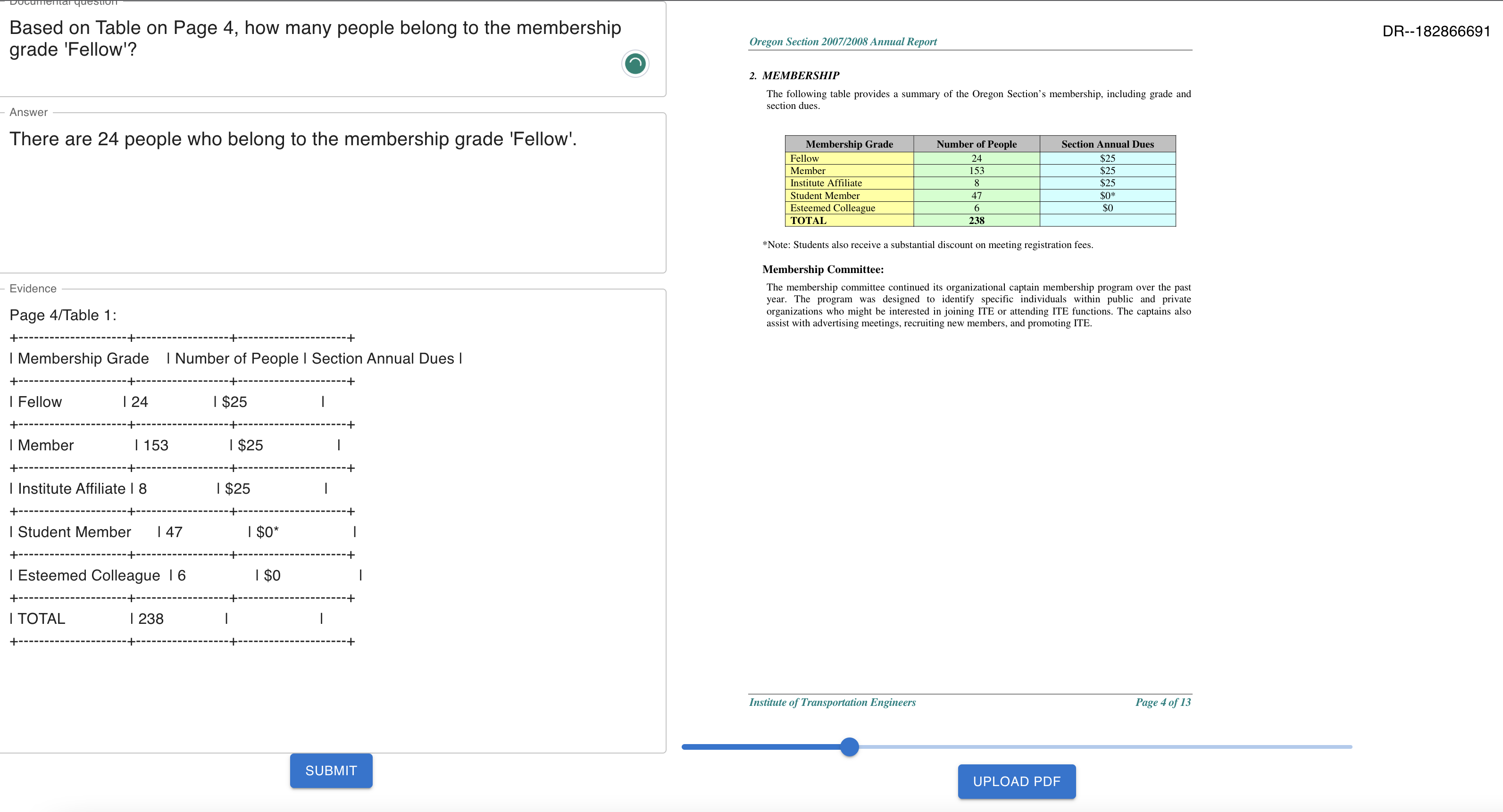}
    \caption{Table QA asking for the number of people belonging to the membership grade `Fellow'. It is shown that ChatGPT can understand table structure in the format of markdown and successfully fetch the number of people belonging to membership `Fellow'.}
    \label{fig-Table1-ex}
\end{figure}

\begin{figure}[htbp!]
    \centering
    \includegraphics[width=\textwidth]{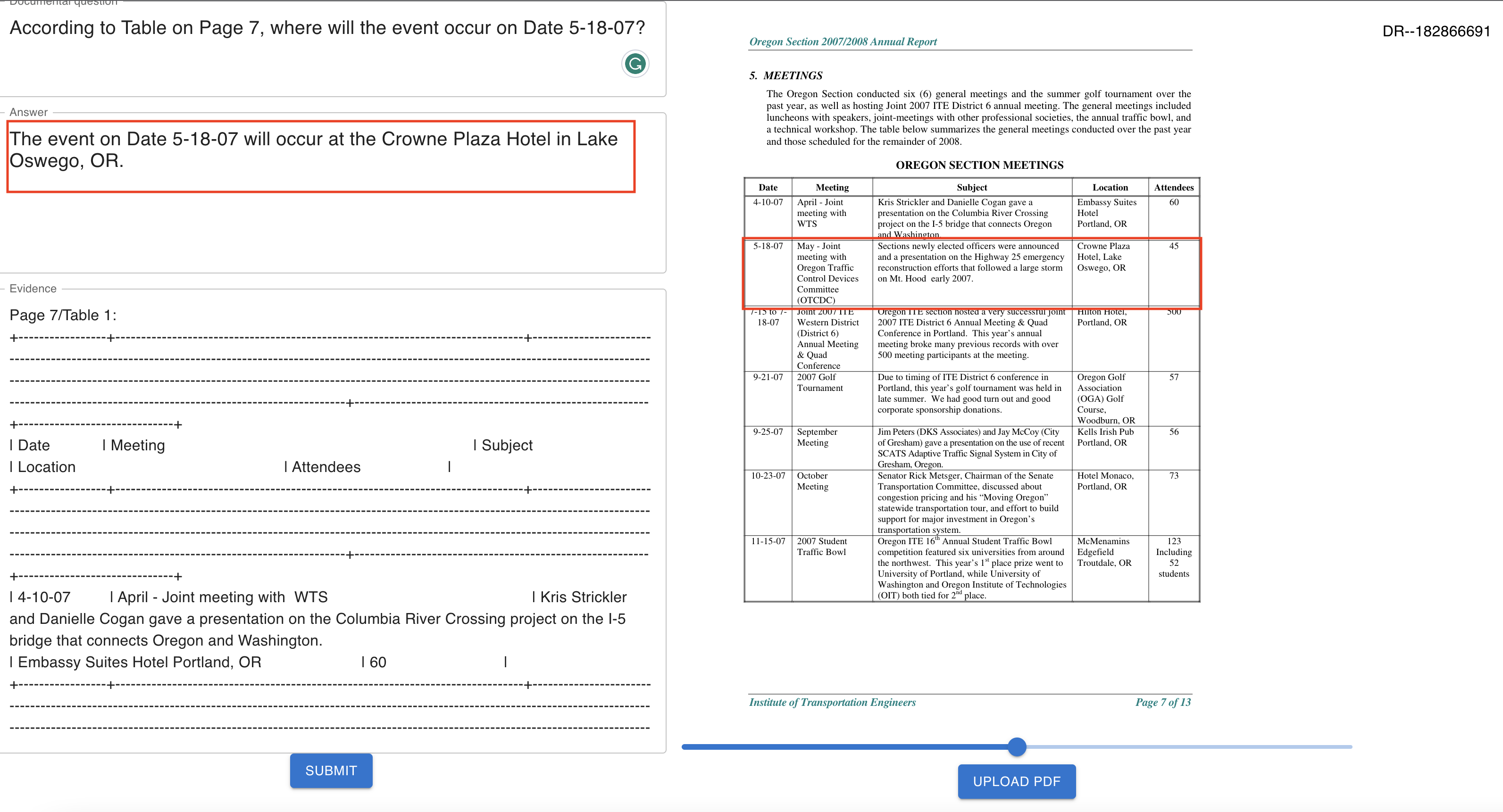}
    \caption{Table QA asking for the place where the event on Date 5-18-07 will occur.}
    \label{fig-Table2-ex}
\end{figure}

\begin{figure}[htbp!]
    \centering
    \includegraphics[width=\textwidth]{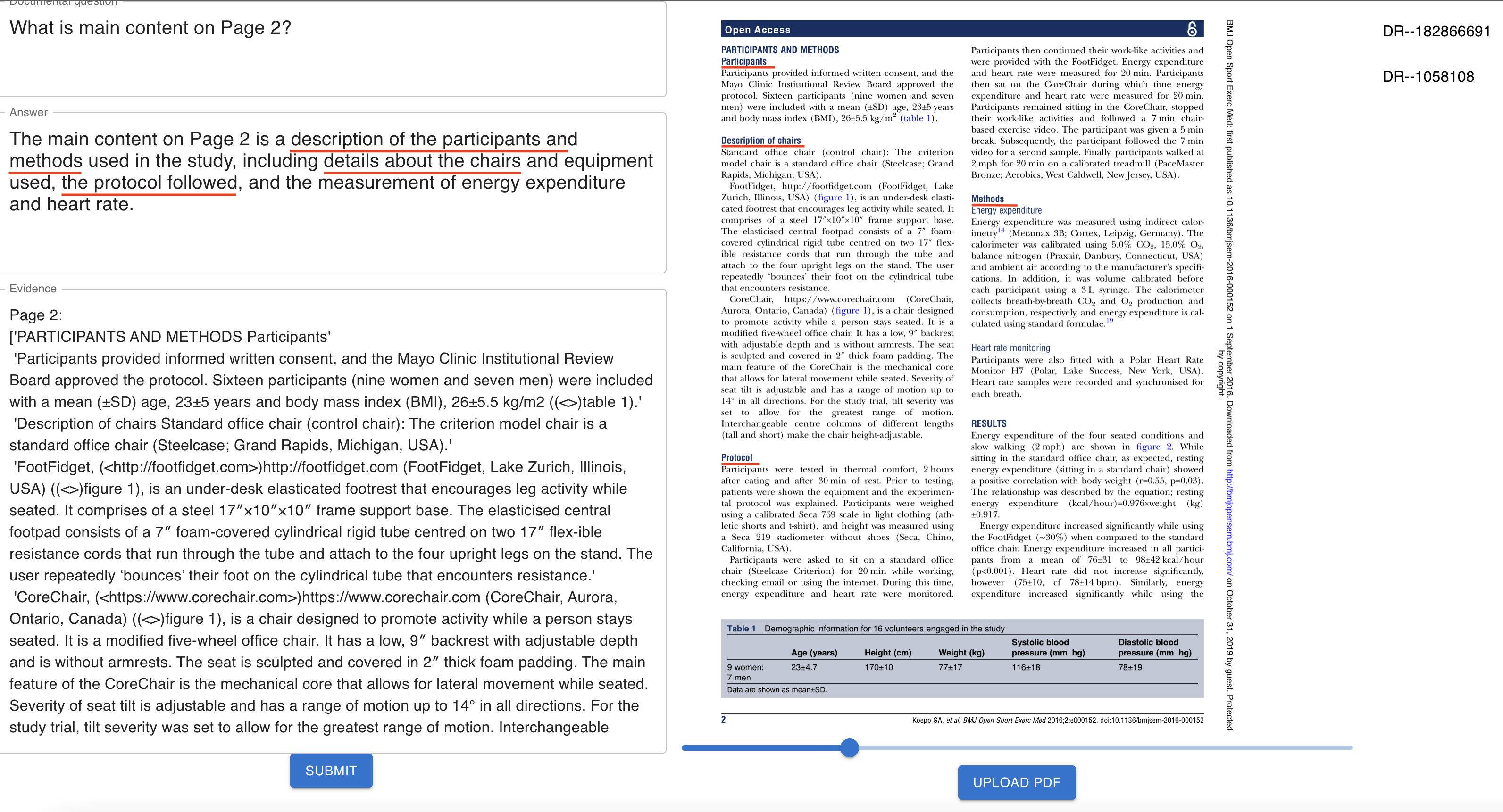}
    \caption{Page QA asking the main content on Page 2. The answer provides a high-level summarization of Page 2, covering the title of each section.}
    \label{fig-Page-ex}
\end{figure}

\begin{figure}[htbp!]
    \centering
    \includegraphics[width=\textwidth]{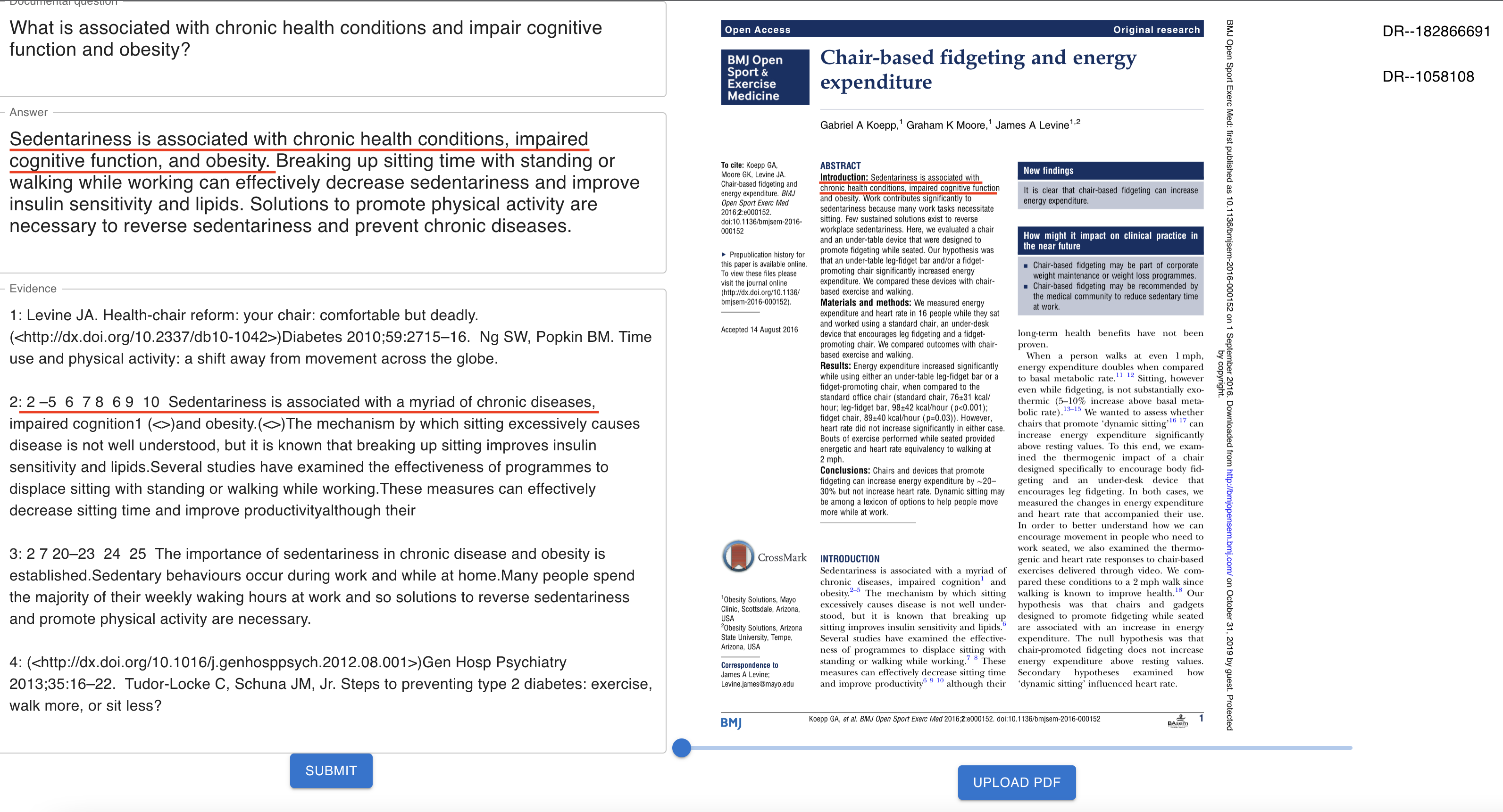}
    \caption{Single Document Content QA asking Sedentariness. The 2$^{\textbf{nd}}$ retrieved sentence includes the answer and corresponds to the first sentence in the abstract of the paper.}
    \label{fig-singledoc-ex}
\end{figure}

\begin{figure}[htbp!]
    \centering
    \includegraphics[width=\textwidth]{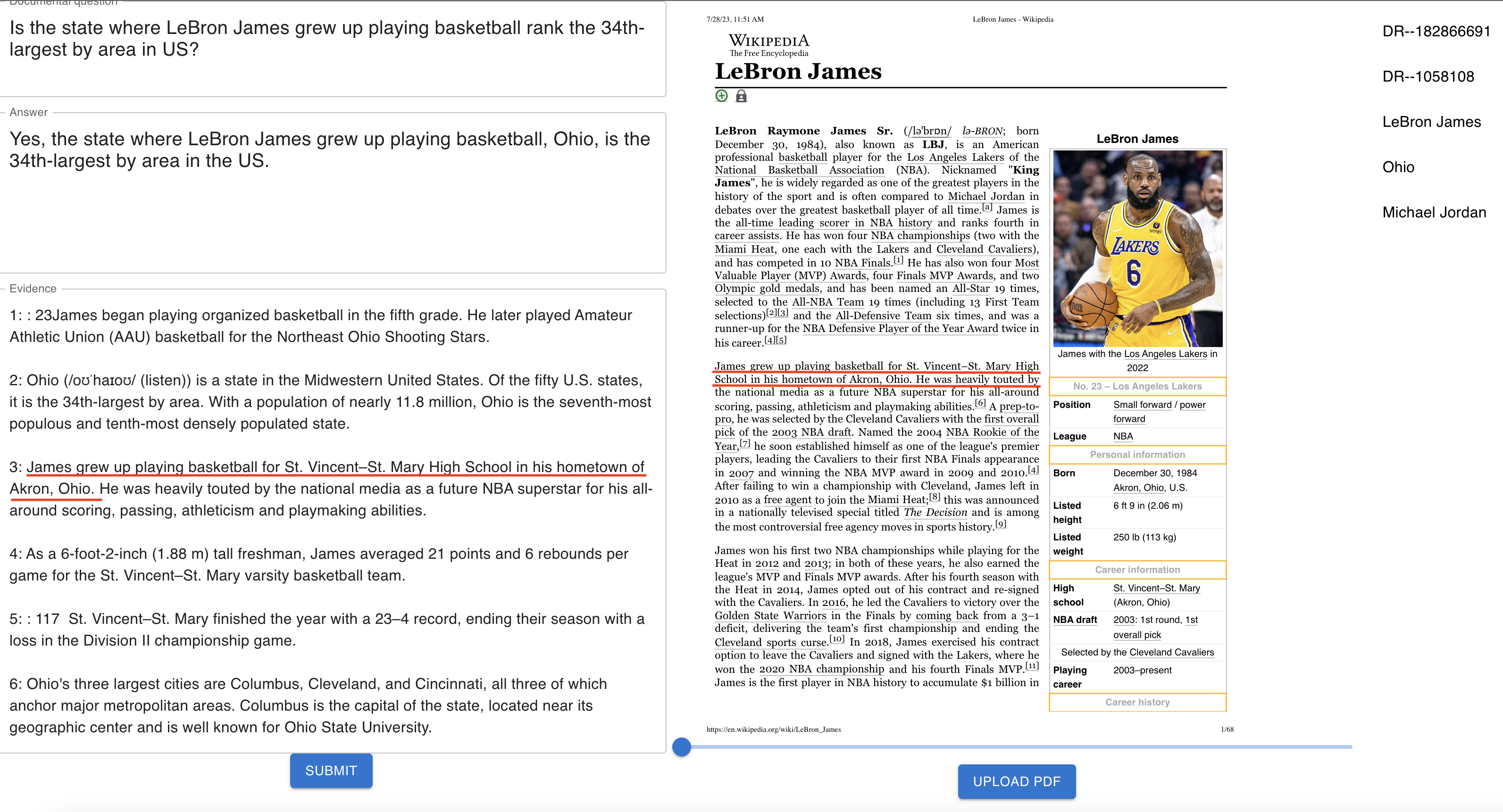}
    \caption{Multi-document Bridging Question asking the information about Lebron James and State Ohio. It requires to first retrieve the sentence stating the state where Lebron James grew up playing basketball.}
    \label{fig-LBJOhio-ex1}
\end{figure}

\begin{figure}[htbp!]
    \centering
    \includegraphics[width=\textwidth]{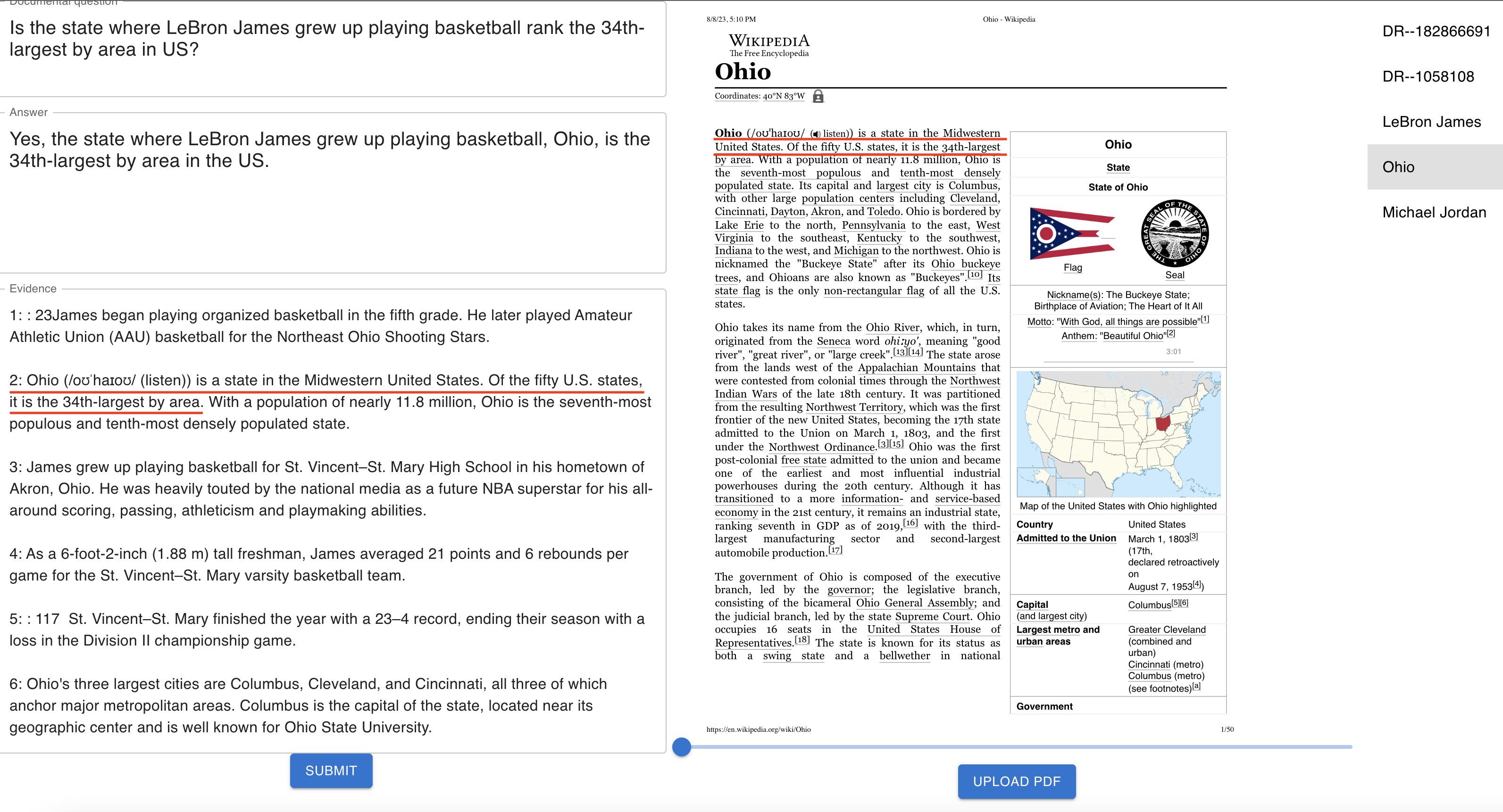}
    \caption{Multi-document Bridging Question asking the information about Lebron James and State Ohio. Then it requires to judge whether the State Ohio ranks the 34th-largest by area in the US.}
    \label{fig-LBJOhio-ex2}
\end{figure}

\begin{figure}[htbp!]
    \centering
    \includegraphics[width=\textwidth]{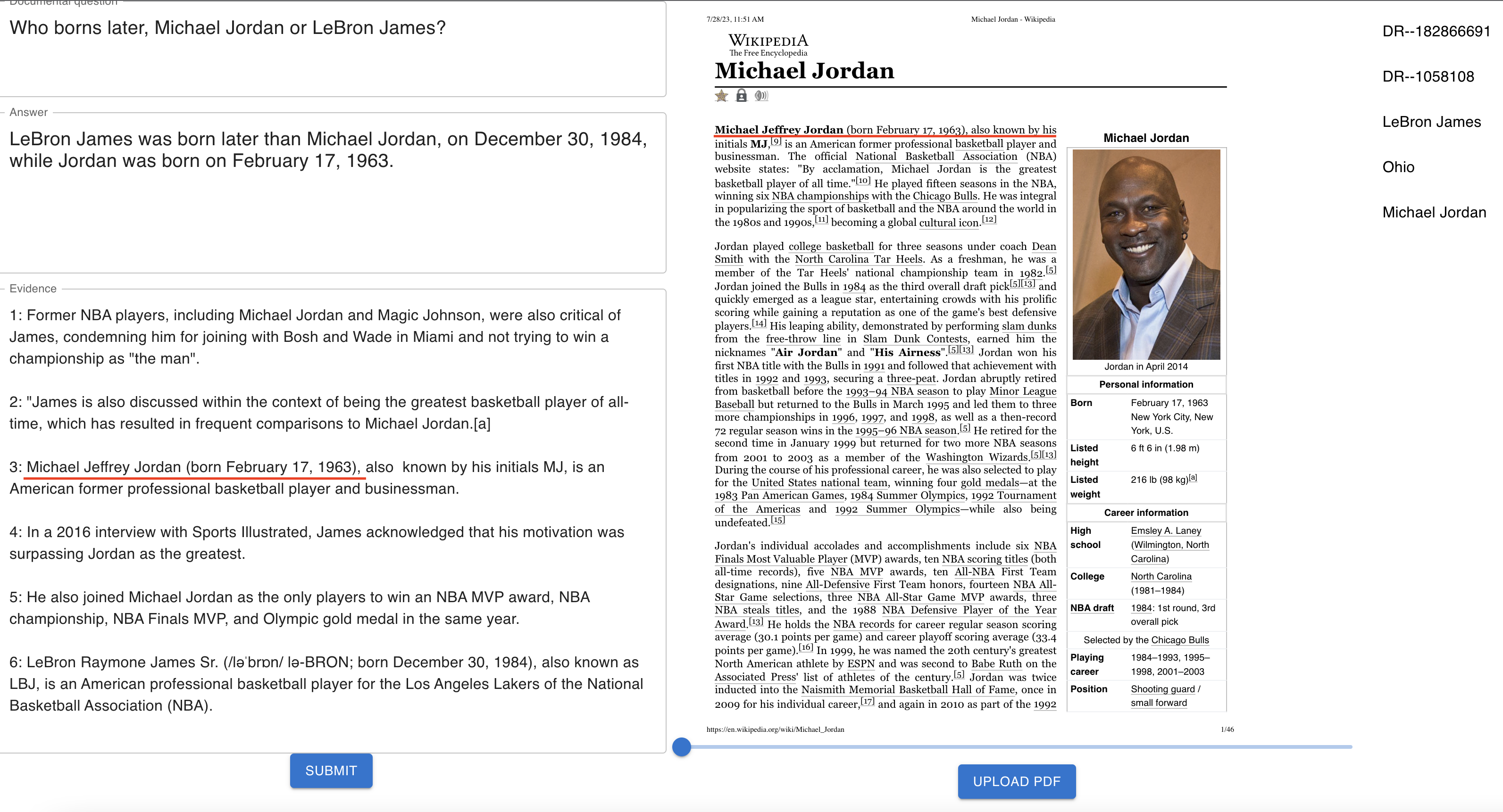}
    \caption{Multi-document Comparing Question comparing Lebron James and Michael Jordan. It requires the birthday information of Lebron and Jordan.}
    \label{fig-MJLBJ-ex1}
\end{figure}

\begin{figure}[htbp!]
    \centering
    \includegraphics[width=\textwidth]{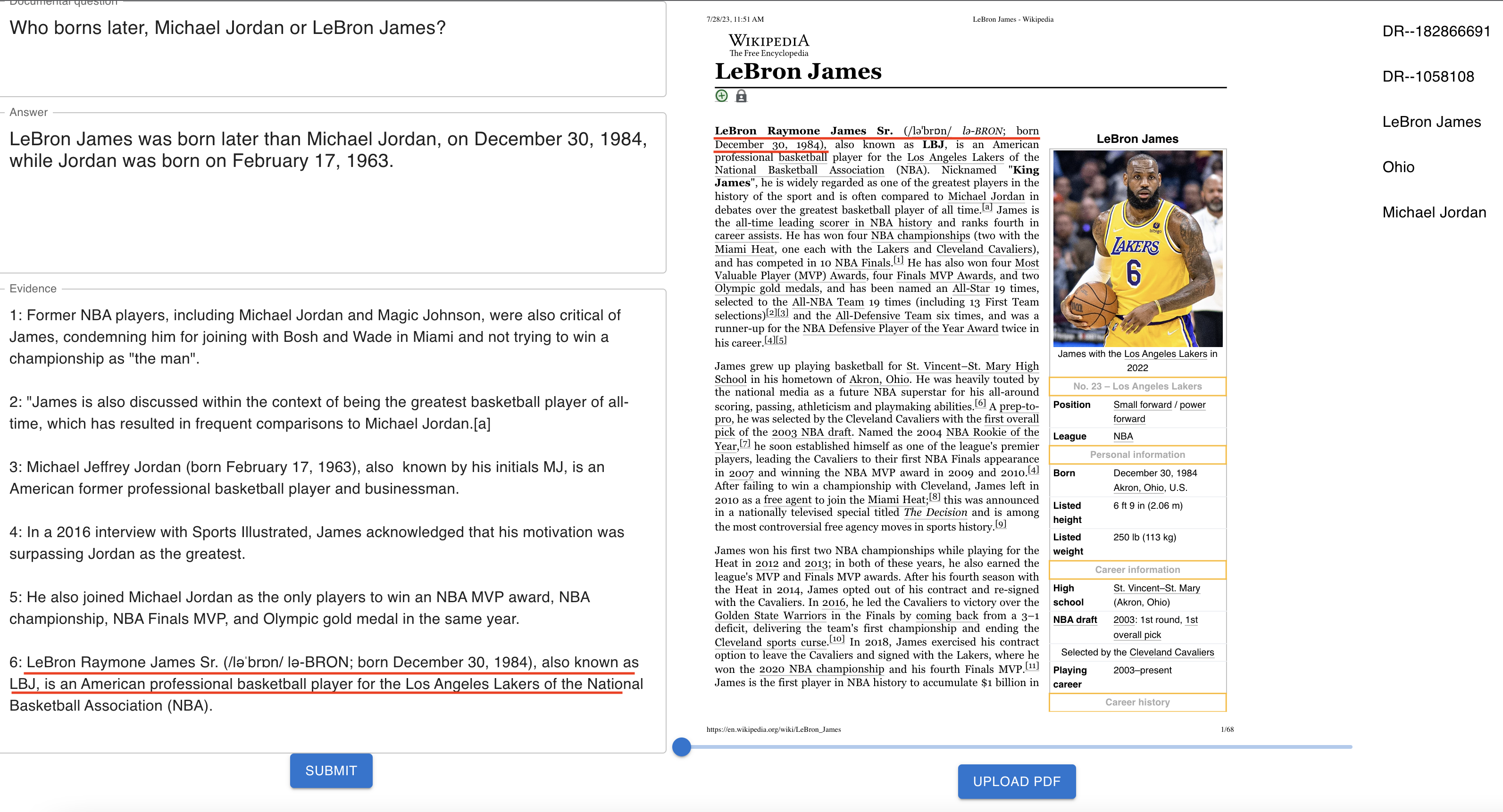}
    \caption{Multi-document Comparing Question comparing Lebron James and Michael Jordan.  It requires the birthday information of Lebron and Jordan.}
    \label{fig-MJLBJ-ex2}
\end{figure}

In this section, we conduct six MD-QA case studies using our self-designed user interface coupled with the proposed method on the backend. Examples include two table-based QA (Figure~\ref{fig-Table1-ex}-\ref{fig-Table2-ex}), one page-based QA (Figure~\ref{fig-Page-ex}), one single-document content-based QA (Figure~\ref{fig-singledoc-ex}) and two multi-document content-based QA (Figure~\ref{fig-LBJOhio-ex1}-\ref{fig-MJLBJ-ex2}). In our designed interface, we can upload documents we are interested in reading and the model on the backend will split each of them into multiple passages. In addition, on the left side, we can ask questions related to the currently uploaded documents. By clicking the button `SUBMIT', the question would be sent to the model on the backend and it retrieves relevant context and arranges them as the prompt to get the answer from ChatGPT. In the figures below, we can see our system can understand the Table/Page questions and also questions requiring knowledge across multiple documents.

\onecolumn

\newpage
\subsection{Visualizing the Reasoning-and-Retrieving Process of LM-guided Graph Traverser}\label{sec-visualreason}

\begin{figure}[htbp!]
    \centering
    \includegraphics[width=\textwidth]{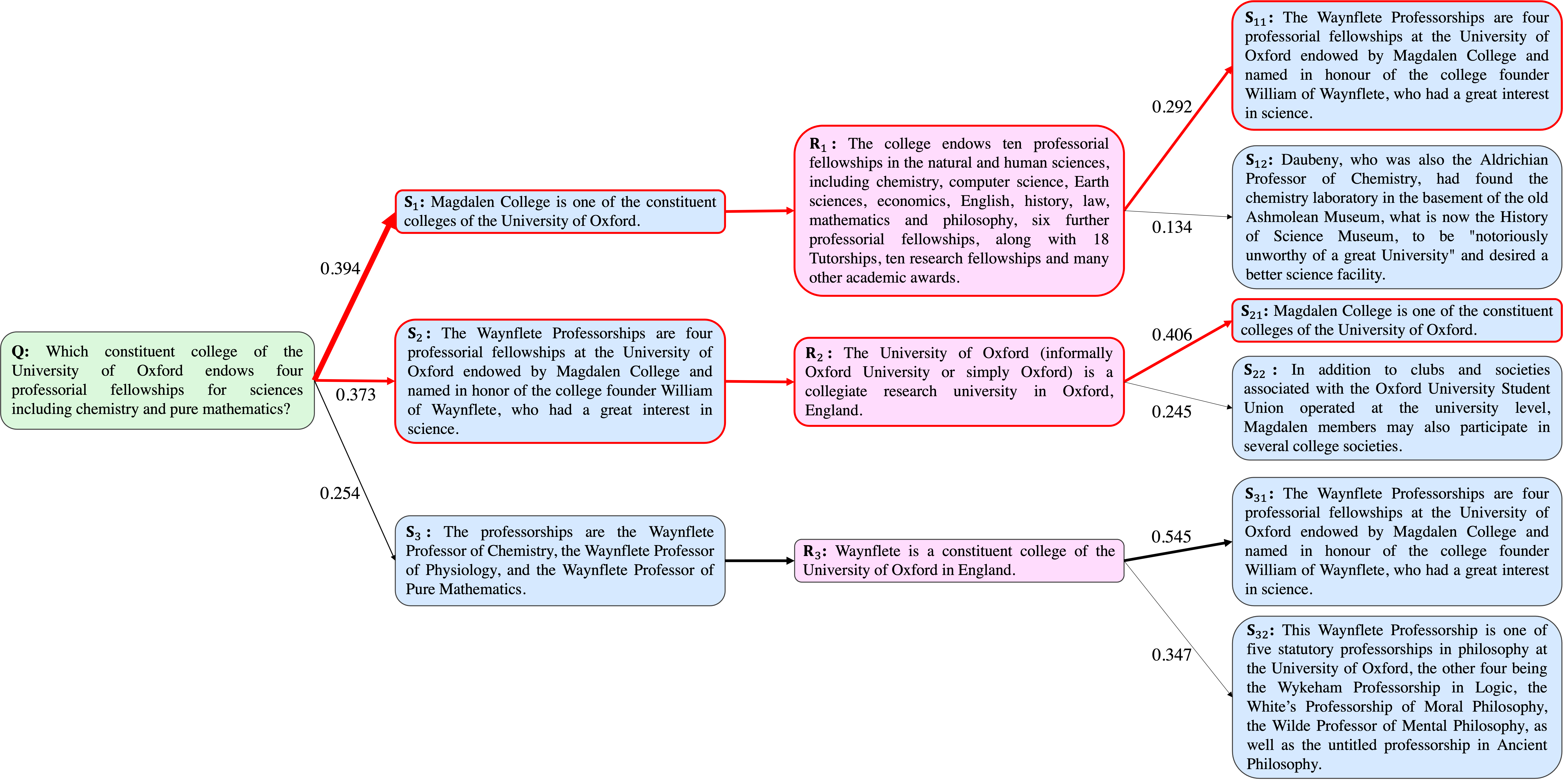}
    \caption{Visualizing the graph traversal over MD-QA-Example 1.}
    \label{fig-visual-ex1}
\end{figure}

\begin{figure}[htbp!]
    \centering
    \includegraphics[width=\textwidth]{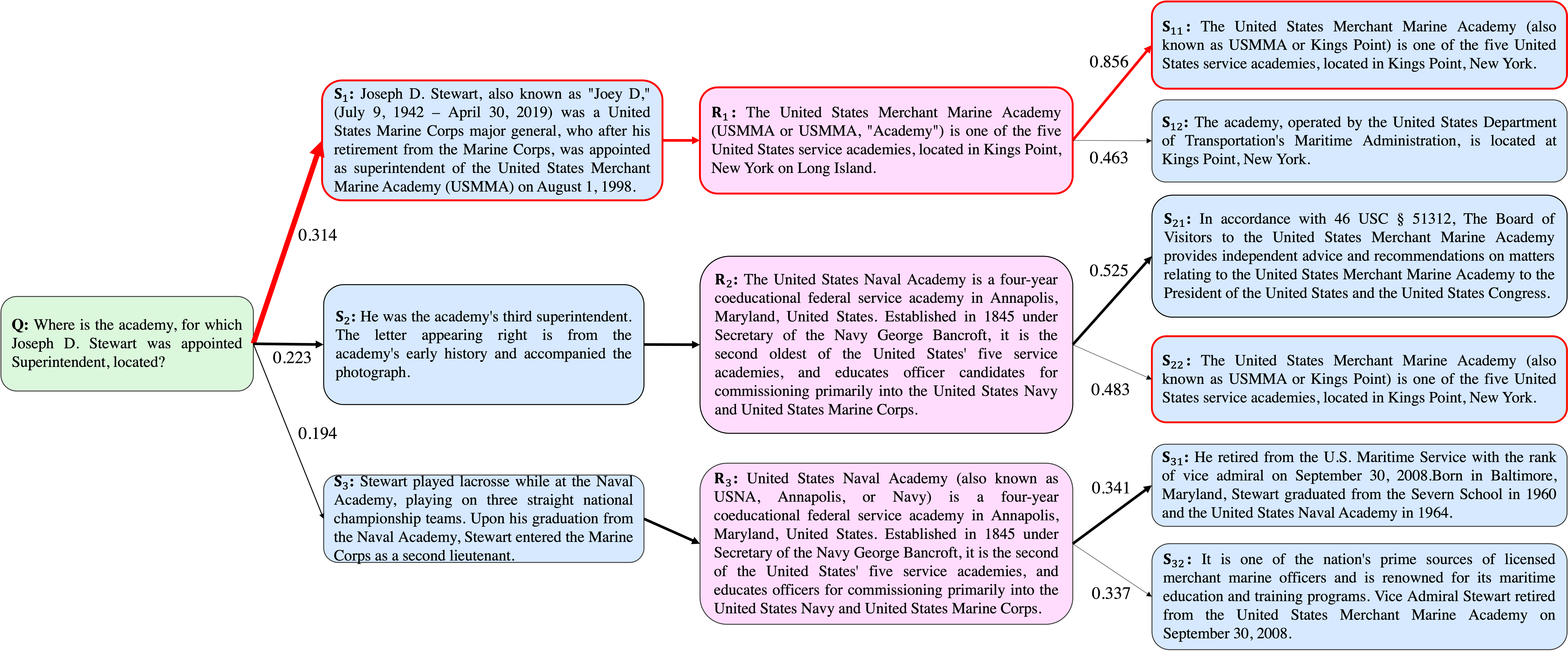}
    \caption{Visualizing the graph traversal over MD-QA-Example 2.}
    \label{fig-visual-ex2}
\end{figure}

\newpage
\subsection{Prompt template used throughout this work}\label{sec-prompt}
\colorlet{shadecolor}{gray!10}
\UseRawInputEncoding
\lstset{breaklines=true, columns=fullflexible, backgroundcolor=\color{shadecolor}}
\begin{lstlisting}[basicstyle=\footnotesize, title={Examples of the Instruction Data for Fine-tuning LLaMA and T5-Large.}, caption={Examples of the Instruction Data for Fine-tuning LLaMA.}, label={list-InstructData}]
Question: Which magazine was started first Arthur's Magazine or First for Women? 
Answer: Arthur's Magazine
Supporting Facts: 
(1) Arthur's Magazine (1844-1846) was an American literary periodical published in Philadelphia in the 19th century.
(2) First for Women is a woman's magazine published by Bauer Media Group in the USA. The magazine was started in 1989.

Instruction: What evidence do we need to answer the question given the current evidence?
Input: Which magazine was started first Arthur's Magazine or First for Women? Arthur's Magazine (1844-1846) was an American literary periodical published in Philadelphia in the 19th century.
Output: First for Women is a woman's magazine published by Bauer Media Group in the USA. The magazine was started in 1989.
===================================================================================================
Question: In what year was the creator of the current arrangement of Simpson's Theme born?
Answer: March 28, 1941
Supporting Facts: 
(1) The theme was re-arranged during season 2, and the current arrangement by Alf Clausen was introduced at the beginning of season 3.
(2) Alf Heiberg Clausen (born March 28, 1941) is an American film and television composer.

Instruction: What evidence do we need to answer the question given the current evidence?
Input: In what year was the creator of the current arrangement of Simpson's Theme born? The theme was re-arranged during season 2, and the current arrangement by Alf Clausen was introduced at beginning of season 3.
Output: Alf Heiberg Clausen (born March 28, 1941) is an American film and television composer.
\end{lstlisting}

\colorlet{shadecolor}{gray!10}
\UseRawInputEncoding
\lstset{breaklines=true, columns=fullflexible, backgroundcolor=\color{shadecolor}}
\begin{lstlisting}[basicstyle=\footnotesize, title={Example of the Prompt for QA without Retrieved Contexts.}, caption={Example of the Prompt for QA without Retrieved Contexts.}, label={list-Prompt1}]
Given the following question, create a final answer to the question.
=========
QUESTION: What is the birthday of this Anglo-Irish actress, courtesan, and mistress, who was the mother to the illegitimate daughter of King William IV?
=========
ANSWER: Please answer in less than 6 words.
\end{lstlisting}

\colorlet{shadecolor}{gray!10}
\UseRawInputEncoding
\lstset{breaklines=true, columns=fullflexible, backgroundcolor=\color{shadecolor}}
\begin{lstlisting}[basicstyle=\footnotesize, title={Example of the Prompt for QA with Retrieved Contexts.}, caption={Example of the Prompt for QA with Retrieved Contexts.}, label={list-Prompt2}]
Given the following question and contexts, create a final answer to the question.
=========
QUESTION: During which years was the model of car, featured on the cover of Earth's "Pentastar: In the Style of Demons" manufactured?
=========
CONTEXT:
1: Pentastar: In the Style of Demons is the third full-length studio album by the drone doom band Earth.
2: In 1957, he published The Interpersonal Diagnosis of Personality, which the Annual Review of Psychology called the "most important book on psychotherapy of the year".
3: During the evanescent heyday of the cyberdelic counterculture, he served as a consultant to Billy Idol in the production of the 1993 album Cyberpunk.
4: During the development of the Barracuda, one of the worst-kept secrets was Ford's plan to introduce a new sporty compact car based on the inexpensive Falcon chassis and running gear (which was eventually released as the Mustang in mid-model year 1964); the extent of the other changes was not known.
5: "Peace in Mississippi" is a cover of the Jimi Hendrix song. The original vinyl release of the album has an alternative take of "Peace in Mississippi".
6: A 1975 Barracuda had been planned before the end of the 1970-74 model cycle.
7: In the spring of 2021, when the third wave of the coronavirus epidemic arrived, Váradi called their airline one of the "rare rays of hope" for investors.
8: During this time the first U.S. Federal auto safety standards were phased in, and Chrysler's response a requirement for side-marker lights distinguishes each model year of the second-generation Barracuda:As the pony-car class became established and competition increased, Plymouth began to revise the Barracuda's engine options.
9: The Barracuda sold for a base price of US$2,512 ($24,000 today).The 1964 model year was the first for the Barracuda and also the last year for push-button control of the optional Torqueflite automatic transmission.
10: In the words of symbolist poet Stéphane Mallarmé:Languages are imperfect because multiple; the supreme language is missing...no one can utter words which would bear the miraculous stamp of Truth Herself Incarnate...how impossible it is for language to express things...in the Poet's hands...by the consistent virtue and necessity of an art which lives on fiction, it achieves its full efficacy.
11: In France, the heart of the Decadent movement was during the 1880s and 1890s, the time of fin de siècle, or end-of-the-century gloom.
12: Pentastar: In the Style of Demons is the third full-length studio album by the drone doom band Earth, released in 1996. It has a more rock-oriented sound than their earlier drone doom work, although in a very minimalist style.
13: The game was a rematch of the previous year's Russell Athletic Bowl, which Clemson won 40–6.The two participants for the game were two of the semifinalists which were the Clemson Tigers and Oklahoma Sooners.
14: The effect of the war on Ernst was devastating; in his autobiography, he wrote of his time in the army thus: "On the first of August 1914 M[ax].E[rnst]. died. He was resurrected on the eleventh of November 1918".
15: Plymouth's executives had wanted to name the new model Panda, an idea unpopular with its designers. In the end, John Samsen's suggestion of Barracuda prevailed. Based on Chrysler's A-body, the Barracuda debuted in fastback form on April 1, 1964.
16: The Scapigliati (literally meaning "unkempt" or "disheveled") were a group of writers and poets who shared a sentiment of intolerance for the suffocating intellectual atmosphere between the late Risorgimento (1860s) and the early years of unified Italy (1870s).
17: Recurrent themes in his literary works include the supremacy of the individual, the cult of beauty, exaggerated sophistication, the glorification of machines, the fusion of man with nature, and the exalted vitality coexisting with the triumph of death.
18: Disc brakes and factory-installed air conditioning became available after the start of the 1965 model year. For the 1966 model year, the Barracuda received new taillamps, new front sheet metal, and a new instrument panel.
19: "Perhaps the worst failing of the book is the omission of any kind of proof for the validity and reliability of the diagnostic system," Eysenck wrote.
20: Based on stretched underpinnings of the rear-drive Alfa Romeo Giulia, it was rumored to be powered by a turbocharged V6 and arrive within the 2019 model year.
21: Their investments are in fleet development and the construction of airports, the first of which will be opened in Brasov.
22: He broke the hill record and this innovation was widely copied in the years to come.[citation needed]Mays made his mark on the track in such events as the 1935 German Grand Prix (scene of a famous victory of Tazio Nuvolari), sharing his ERA with Ernst von Delius.
23: There is still a question about the truth of the disclosure. In the 1968 Dragnet episode "The Big Prophet", Liam Sullivan played Brother William Bentley, leader of the Temple of the Expanded Mind, a thinly fictionalized Leary.
24: The Belgian Félicien Rops was instrumental in the development of this early stage of the Decadent movement. A friend of Baudelaire, he was a frequent illustrator of Baudelaire's writing, at the request of the author himself.
25: After taking responsibility for the controlled substance, Leary was convicted of possession under the Marihuana Tax Act of 1937 on March 11, 1966, sentenced to 30 years in prison, fined $30,000, and ordered to undergo psychiatric treatment.
26: The general court delegation from Sullivan County is made up of all of the members of the New Hampshire House of Representatives from the county. In total, there are 13 members from 11 different districts.
27: Both teams then exchanged field goals, which brought the score to 16-10 in favor of Clemson. With 2:17 remaining, Oklahoma drove down the length of the field to score a touchdown, which gave the Sooners a one-point lead.
28: The average household size was 2.41 and the average family size was 2.88.23.90% of the population were under the age of 18, 6.40% from 18 to 24, 28.00% from 25 to 44, 25.90% from 45 to 64, and 15.80% who were 65 years of age or older.
29: The band announced the release of a deluxe version of the album "How It Feels To Be Lost", which came out on August 21, 2020. On June 2, 2021, the band released the single "Bloody Knuckles" from their upcoming album.
30: The 82nd Orange Bowl was a College Football Playoff semifinal with the winner of the game competing against the winner of the 2015 Cotton Bowl: Alabama Crimson Tide football in the 2016 College Football Playoff National Championship, which took place at the University of Phoenix Stadium in Glendale, Arizona.
=========
QUESTION: During which years was the model of car, featured on the cover of Earth's "Pentastar: In the Style of Demons" manufactured?
=========
ANSWER: Please answer in less than 6 words.
\end{lstlisting}

\newpage
\colorlet{shadecolor}{gray!10}
\UseRawInputEncoding
\lstset{breaklines=true, columns=fullflexible, backgroundcolor=\color{shadecolor}}
\begin{lstlisting}[basicstyle=\footnotesize, title={Example of the Prompt for QA with Retrieved Contexts for MDR, KGP-T5, KGP-LLaMA and KGP-MDR.}, caption={Example of the Prompt for QA with Retrieved Contexts for MDR, KGP-T5, KGP-LLaMA and KGP-MDR.}, label={list-Prompt3}]
Given the following question and contexts, create a final answer to the question.
=========
QUESTION: Anthony Avent played basketball for a High School that is located in a city approximately 8 mi west of where?
=========
CONTEXT:
1:  Newark is the second largest city in the New York metropolitan area, located approximately 8 mi west of lower Manhattan.\n Prior to Seton Hall, Avent played at Malcolm X Shabazz High School in Newark, New Jersey.

2:  Newark is the second largest city in the New York metropolitan area, located approximately 8 mi west of lower Manhattan.\n The United States District Court for the District of New Jersey is also located in the city.

3:  Newark is the second largest city in the New York metropolitan area, located approximately 8 mi west of lower Manhattan.\n Near Market Street and includes a dormitory for boarding students; and Saint Vincent Academy which is an all-girls Roman Catholic high school founded and sponsored by the Sisters of Charity of Saint Elizabeth and operated continuously since 1869.Link Community School is a non-denominational coeducational day school that serves approximately 128 students in seventh and eighth grades.

4:  Prior to Seton Hall, Avent played at Malcolm X Shabazz High School in Newark, New Jersey.\n Newark is the second largest city in the New York metropolitan area, located approximately 8 mi west of lower Manhattan.

5:  Prior to Seton Hall, Avent played at Malcolm X Shabazz High School in Newark, New Jersey.\n The United States District Court for the District of New Jersey is also located in the city.

6:  Prior to Seton Hall, Avent played at Malcolm X Shabazz High School in Newark, New Jersey.\n On Newark Bay, it is run by the Port Authority of New York and New Jersey and serves as the principal container ship facility for goods entering and leaving the New York metropolitan area and the northeastern quadrant of North America.

7: He played collegiately at Seton Hall University where he played in the 1989 NCAA championship game. Prior to Seton Hall, Avent played at Malcolm X Shabazz High School in Newark, New Jersey.\n Prior to Seton Hall, Avent played at Malcolm X Shabazz High School in Newark, New Jersey.

8: He played collegiately at Seton Hall University where he played in the 1989 NCAA championship game. Prior to Seton Hall, Avent played at Malcolm X Shabazz High School in Newark, New Jersey.\n The United States District Court for the District of New Jersey is also located in the city.

9: He played collegiately at Seton Hall University where he played in the 1989 NCAA championship game. Prior to Seton Hall, Avent played at Malcolm X Shabazz High School in Newark, New Jersey.\n As of the 2020–21 school year, the district, comprises 65 schools, had an enrollment of 40,423 students and 2,886.5 classroom teachers (on an FTE basis), for a student–teacher ratio of 14.0:1.Science Park High School, which was the 69th-ranked public high school in New Jersey out of 322 schools statewide, in New Jersey Monthly magazine's September 2010 cover story on the state's "Top Public High Schools", after being ranked 50th in 2008 out of 316 schools.

10: Anthony Avent (born October 18, 1969) is an American former professional basketball player who was selected by the Atlanta Hawks in the first round (15th pick overall) of the 1991 NBA draft.\n Newark is the second largest city in the New York metropolitan area, located approximately 8 mi west of lower Manhattan.

11: Anthony Avent (born October 18, 1969) is an American former professional basketball player who was selected by the Atlanta Hawks in the first round (15th pick overall) of the 1991 NBA draft.\n The United States District Court for the District of New Jersey is also located in the city.

12: Anthony Avent (born October 18, 1969) is an American former professional basketball player who was selected by the Atlanta Hawks in the first round (15th pick overall) of the 1991 NBA draft.\n Atlanta United 1, New York Red Bulls 2 The first game in Atlanta United history was played before a sellout crowd of 55,297.

13: Anthony Avent (born October 18, 1969) is a retired American professional basketball player who was selected by the Atlanta Hawks in the first round (15th pick overall) of the 1991 NBA Draft.\n The total school enrollment in Newark was 77,097 in the 2013–2017 ACS, with nursery and preschool enrollment of 7,432, elementary/high school (K–12) enrollment of 49,532, and total college/graduate school enrollment of 20,133. The Newark Public Schools, a state-operated school district, is the largest school system in New Jersey.

14: Anthony Avent (born October 18, 1969) is a retired American professional basketball player who was selected by the Atlanta Hawks in the first round (15th pick overall) of the 1991 NBA Draft.\n As of the 2020–21 school year, the district, comprises 65 schools, had an enrollment of 40,423 students and 2,886.5 classroom teachers (on an FTE basis), for a student–teacher ratio of 14.0:1.Science Park High School, which was the 69th-ranked public high school in New Jersey out of 322 schools statewide, in New Jersey Monthly magazine's September 2010 cover story on the state's "Top Public High Schools", after being ranked 50th in 2008 out of 316 schools.

15: Anthony Avent (born October 18, 1969) is a retired American professional basketball player who was selected by the Atlanta Hawks in the first round (15th pick overall) of the 1991 NBA Draft.\n In the 2013--2017 American Community Survey, 13.6% of Newark residents ages 25 and over had never attended high school and 12.5% didn't graduate from high school, while 74.1% had graduated from high school, including the 14.4% who had earned a bachelor's degree or higher.
=========
QUESTION: Anthony Avent played basketball for a High School that is located in a city approximately 8 mi west of where?
=========
ANSWER: Please answer in less than 6 words.
\end{lstlisting}

\colorlet{shadecolor}{gray!10}
\UseRawInputEncoding
\lstset{breaklines=true, columns=fullflexible, backgroundcolor=\color{shadecolor}}
\begin{lstlisting}[basicstyle=\footnotesize, title={Example of the Prompt for Grading QA.}, caption={Example of the Prompt for Grading QA.}, label={list-Prompt4}]
You are an expert professor specialized in grading whether the prediction to the question is correct or not according to the real answer.
==================
For example:
==================
Question: What company owns the property of Marvel Comics?
Answer: The Walt Disney Company
Prediction: The Walt Disney Company
Return: 1
==================
Question: Which constituent college of the University of Oxford endows four professorial fellowships for sciences including chemistry and pure mathematics?
Answer: Magdalen College
Prediction: Magdalen College.
Return: 1
==================
Question: Which year was Marvel started?
Answer: 1939
Prediction: 1200
Return: 0
==================
You are grading the following question:
Question: Anthony Avent played basketball for a High School that is located in a city approximately 8 mi west of where?
Answer: lower Manhattan
Prediction: Newark
If the prediction is correct according to the answer, return 1. Otherwise, return 0.
Return: your reply can only be one number '0' or '1'
\end{lstlisting}